\documentclass[review, sort&compress]{elsarticle}

\usepackage{fullpage}
\usepackage{amssymb}
\usepackage{array}  
\usepackage{pifont}
\usepackage{bbding}
\usepackage{graphicx}
\usepackage{subfig}
\usepackage{enumerate}
\usepackage{amsmath,mathtools}
\usepackage{dsfont}
\usepackage{amsfonts,bm}
\usepackage{todonotes}
\usepackage[commentmarkup=footnote]{changes}
\definechangesauthor[name={Xiao}, color=purple]{hx}
\definechangesauthor[name={Zhou}, color=blue]{xz}
\usepackage{stmaryrd}
\usepackage{siunitx}
\usepackage{caption}
\usepackage{array}
\usepackage{booktabs}
\usepackage{nomencl}
\usepackage{lineno}
\usepackage{url}
\usepackage{soul}
\usepackage{natbib}
\usepackage[strings]{underscore}
\usepackage{tikz}
\usepackage{siunitx}
\usepackage{multirow}
\usepackage{tabularx}
\usepackage{makecell}
\newcommand{\bTheta}{\bm{\Theta}}

\newcommand{\cD}{\mathcal{D}}

\newcolumntype{C}[1]{>{\centering\arraybackslash}p{#1}}

\renewcommand\nomgroup[1]{%
  \item[\itshape
  \ifstrequal{#1}{A}{Symbols}{%
  \ifstrequal{#1}{B}{Roman Letters}{%
  \ifstrequal{#1}{C}{Greek Letters}{%
  \ifstrequal{#1}{D}{Abbreviations}{}}}}%
]}

\makeatletter
\renewcommand\paragraph{%
  \@startsection{paragraph}{4}{\z@}%
  {3.25ex \@plus1ex \@minus.2ex}%
  {-1em}%
  {\normalfont\normalsize\bfseries}%
}
\makeatother

\usepackage{xpatch}
\xpatchcmd{\thenomenclature}{\section*{\nomname}
}{}{\typeout{Success}}{\typeout{Failure}}
\makenomenclature
\RequirePackage{ifthen}
\printnomenclature[0.8cm]

\usepackage[commentmarkup=footnote]{changes}
\definechangesauthor[name={Reviewer 1}, color = red]{R1}
\definechangesauthor[name={Reviewer 2}, color = brown]{R2}
\definechangesauthor[name={Author}, color = orange]{Author}
\usepackage{xcolor}

\graphicspath{{./figs/}}

\begin{document}

\begin{frontmatter}

\title{\textbf{BI-EqNO}: Generalized Approximate \textbf{B}ayesian \textbf{I}nference with an \textbf{Eq}uivariant \textbf{N}eural \textbf{O}perator Framework}

\author[vt]{Xu-Hui Zhou}%
\author[st]{Zhuo-Ran Liu}
\author[st]{Heng Xiao\corref{cor1}}%
\ead{heng.xiao@simtech.uni-stuttgart.de}

\cortext[cor1]{Corresponding author.}

\affiliation[vt]{organization={Kevin T. Crofton Department of Aerospace and Ocean Engineering, Virginia Tech},
            city={Blacksburg},
            postcode={24060}, 
            state={VA},
            country={USA}}
\affiliation[st]{organization={Stuttgart Center for Simulation Science, University of Stuttgart},
            city={Stuttgart},
            postcode={70569}, 
            state={BW},
            country={Germany}}

\begin{abstract}
Bayesian inference provides a robust statistical framework for updating prior beliefs based on new data via Bayes' theorem. However, performing exact Bayesian inference is often computationally infeasible in practical applications, requiring the use of approximate methods. Although these methods have achieved broad success, they continue to face significant challenges in accurately estimating marginal likelihoods. These challenges arise from the rigid, predefined functional structures of deterministic models, such as Gaussian process, and the limitations of small sample sizes in stochastic models like ensemble Kalman method, among other factors.
In this work, we introduce BI-EqNO, an equivariant neural operator framework for generalized approximate Bayesian inference, developed to improve both deterministic and stochastic approaches. BI-EqNO transforms prior distributions into corresponding posteriors conditioned on observational data, achieving higher efficiency and accuracy through data-driven training. The framework is flexible, accommodating diverse prior and posterior representations with arbitrary discretizations and varying numbers of observations. Crucially, BI-EqNO's architecture is specifically designed to preserve symmetry in these functional transformations, ensuring (1) permutation equivariance between prior and posterior representations and (2) permutation invariance with respect to the observation data.
We demonstrate the utility of this framework with two examples: (1) BI-EqNO as a generalized Gaussian process (gGP) for regression, and (2) BI-EqNO as an ensemble neural filter (EnNF) for sequential data assimilation. Our results indicate that gGP outperforms the traditional Gaussian process by offering a more flexible representation of mean and covariance functions. Furthermore, EnNF not only emulates the widely used ensemble Kalman filter but also has the potential to function as a ``super'' filter, representing a series of ensemble filters, and exhibits superior assimilation performance, particularly in small-ensemble settings.
This study highlights the versatility and effectiveness of the BI-EqNO framework, which enhances approximate Bayesian inference methods through data-driven training, enabling more accurate inferences while reducing computational costs across a wide range of applications.
\end{abstract}


\begin{keyword}
approximate Bayesian inference \sep equivariant neural operator \sep Gaussian process \sep generalized Gaussian process \sep ensemble Kalman filter \sep ensemble neural filter
\end{keyword}

\end{frontmatter}


\section{Introduction}
\label{sec:intro}

Bayesian inference provides a robust statistical framework for integrating prior knowledge with observed data. It has found extensive applications across diverse fields, enhancing predictive capabilities and decision-making processes. For example, in turbulence modeling, Bayesian inference is used to calibrate turbulence models in Reynolds-averaged Navier--Stokes equations with high-fidelity data from experiment measurements and direct numerical simulations~\cite{duraisamy2019turbulence,xiao2016quantifying,edeling2014bayesian,zhang2022ensemble}, thereby improving simulation accuracy for engineering design and optimization. In weather forecasting, it provides the foundation for a wide range of data assimilation methods such as the ensemble Kalman filter, which assimilate data from satellites, radar, and ground observations to refine predictions from numerical weather models~\cite{navon2009data,carrassi2018data,berliner1999bayesian}. Similarly, in medical diagnosis, Bayesian inference integrates clinical knowledge with patient data to enhance diagnostic precision and enable personalized care~\cite{richens2020improving,collins2020bayesian}.
The mathematical process of Bayesian inference involves updating the prior probability distribution $p(\bTheta)$, representing initial beliefs about model parameters $\bTheta$, using new data $\cD$ through the likelihood $p(\cD | \bTheta)$ to obtain the posterior distribution $p(\bTheta | \cD)$:
\begin{equation*}
    p(\bTheta | \cD)=\frac{p(\cD | \bTheta)\; p(\bTheta)}{p(\cD)}= \frac{p(\cD | \bTheta)\; p(\bTheta)}{\int p(\cD | \bTheta)\; p(\bTheta)\; d \bTheta},
\end{equation*}
where $p(\cD)$ is the marginal likelihood serving as a normalization factor. Exact computation of posterior distributions is often computationally challenging, especially for complex models with high-dimensional parameter spaces~\cite{mackay2003information}. This necessitates the use of approximate Bayesian inference methods~\cite{rasmussen2003gaussian}.

\subsection{Approximate Bayesian inference}
Approximate Bayesian inference methods are generally categorized into deterministic and stochastic approaches~\cite{minka2001family}. Deterministic methods, such as maximum a posteriori (MAP) estimation, Laplace approximation, and variational inference, aim to approximate the true posterior with a simpler, computationally tractable distribution. MAP estimation identifies the mode of the posterior distribution, representing the most likely parameter values, without accounting for the uncertainty. Laplace approximation refines this by modeling the posterior as a Gaussian distribution centered at the MAP estimate, with variance derived from the curvature of the log-posterior at the mode. Variational inference uses a broader range of tractable families, such as Gaussians, exponentials, or mixtures, and optimizes the chosen approximation by minimizing the Kullback--Leibler (KL) divergence from the true posterior, allowing for more flexible representations.

Deterministic methods encounter challenges in both selecting suitable distribution families and accurately modeling the selected distributions. Gaussian distributions are frequently chosen for their mathematical tractability in representing diverse phenomena. Once a distribution is selected, it is crucial to model it with appropriate functional structures. For instance, the mean and covariance functions defining a multivariate Gaussian distribution are essential for accurate posterior approximation, especially when data is limited. A common misstep is using an inappropriate covariance function; for example, employing a distance-decaying covariance function to model data with periodic characteristics leads to inaccurate posterior approximations. This underscores the importance of correctly modeling the chosen distribution to achieve reliable Bayesian inference.

Stochastic methods, such as Markov chain Monte Carlo (MCMC) and sequential Monte Carlo (SMC)~\cite{murphy2022probabilistic}, offer greater flexibility since they do not require predefined distribution families. This adaptability makes them particularly effective in managing complex and high-dimensional posterior landscapes.
MCMC methods generate a Markov chain to sample from the target distribution at equilibrium, theoretically providing a comprehensive representation of the posterior distribution. Conversely, SMC methods, such as particle filters and ensemble-based filters~\cite{murphy2022probabilistic}, are designed for real-time data assimilation, where prior estimates are iteratively updated with new sequential observations.
Particle filters represent the posterior distribution using a set of samples that are propagated and reweighted according to observed data, facilitating a flexible, non-parametric approach to handling nonlinear and non-Gaussian process~\cite{gordon1993novel}. The ensemble Kalman filter employs an ensemble of state samples to efficiently approximate the mean and covariance of the posterior, making it particularly suitable for large-scale systems with linear or mildly nonlinear dynamics~\cite{evensen1994sequential,zhou2023inference,pensoneault2023ensemble}.

Despite their flexibility and effectiveness, stochastic methods have a notable limitation in that they often require a sufficiently large sample size to accurately approximate the posterior distribution.
For example, in SMC methods, a large sample size is necessary to capture the full range of system uncertainties, particularly for large-scale systems like atmosphere and ocean. This results in substantial computational intensity, as propagating numerous samples through numerical simulations demands significant processing power and memory. For instance, the European Centre for Medium-Range Weather Forecasts uses an ensemble of 51 samples for weather forecasting, with each run taking approximately 10 petaflop/s of processing power and producing several terabytes of data daily~\cite{bauer2015quiet}. Similarly, ocean modeling using the HYbrid Coordinate Ocean Model with a 1/12$^\circ$ resolution can require tens of thousands of core hours per simulation day~\cite{chassignet2009us}. 
When computational resources are limited, it is crucial to explore more efficient simulation methods and advance filtering techniques. Approaches such as reduced order modeling and surrogate modeling provide viable alternatives to traditional numerical simulations, enabling large-ensemble data assimilation with substantially reduced computational costs~\cite{chattopadhyay2023deep,xia2024characterization}. On the other hand, to improve filtering performance, techniques like localization and inflation are commonly employed to alleviate the challenges posed by low-rank approximations with limited samples in ensemble-based methods~\cite{evensen2022data}. Despite their effectiveness, these methods often require significant problem-specific tuning and may still have limitations in certain scenarios. As a result, developing a more flexible and universally applicable filtering approach that maintains accuracy across a wide range of conditions remains a critical area of ongoing research.

\subsection{Neural networks for approximate Bayesian inference}
Recent advancements in machine learning offer new approaches to addressing the challenges faced by both deterministic and stochastic methods. In deterministic methods, neural networks offer a powerful and flexible tool for modeling probability distributions. Enhancements to Gaussian processes for regression problems demonstrate this utility~\cite{damianou2013deep, lee2017deep, garnelo2018conditional}. Among these, neural processes have emerged as an effective and scalable approach~\cite{garnelo2018conditional}. They employ an encoder-decoder architecture where the encoder maps observation data into a latent space that captures uncertainty and variability, and the decoder maps this latent representation back to prediction values, conditioned on the prediction points. Trained on observation data, the neural processes demonstrate superior predictive capability and reduced computational complexity compared to Gaussian processes. However, they treat each target point individually without considering their covariance, thus neglecting the critical aspect of uncertainty inherent in Gaussian processes.

In stochastic methods, neural networks can provide a more accurate and flexible scheme for updating prior samples based on observations. For example, a U-Net-based neural network has been developed to enhance the ensemble Kalman filter by considering a nonlinear mapping between analysis increments and innovations~\cite{zhang2024novel}. This surrogate filter shows advantages in systems with non-Gaussian state distributions and under small ensemble conditions. However, as it updates each sample individually without considering ensemble statistics, it may not fully leverage the benefits of ensemble-based methods. Some studies retain the ensemble Kalman filter as an essential component of the filtering process and augment it with a trained neural-network corrector~\cite{fan2021combining, brajard2020combining}, resulting in a two-step assimilation process.

Despite the improvements from leveraging neural networks, aforementioned studies often focus on specific application scenarios from a discrete perspective, limiting their broader applicability to real-world problems with diverse prior and posterior representations. Mathematically, we can view the prior-to-posterior transformation as a general functional mapping, which shall be applicable across various representations of both distributions. Recent developments in operator learning provide novel methodologies for handling such functional transformations. Neural operators, initially proposed for the supervised learning of solution operators in partial differential equations (PDEs)~\cite{lu2021learning,li2020fourier,kovachki2023neural}, distinguish themselves from standard neural networks by learning mappings between infinite-dimensional function spaces and adapting to different input discretizations. This adaptability allows them to handle various input forms and scales, making them highly versatile and particularly advantageous in applications requiring high-dimensional functional mappings, such as fluid dynamics and climate modeling. Neural operators have demonstrated success across a variety of fields including constitutive modeling~\cite{zhou2022frame,han2023equivariant}, aerodynamic shape optimization~\cite{li2024geometry,elrefaie2024drivaernet++}, and advancing diverse scientific simulations~\cite{zhou2023neural,azizzadenesheli2024neural,zafar2021frame,chen2024hybrid}.

\begin{figure}[!htb]
\centering
\includegraphics[width=0.75\textwidth]{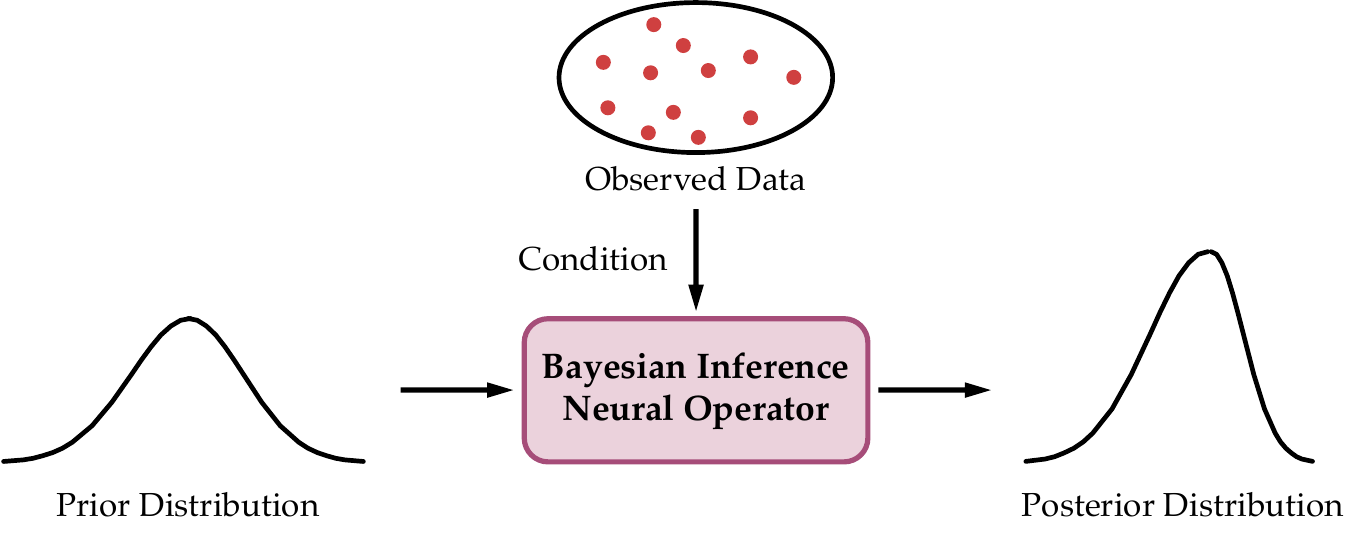}
  \caption{Schematic illustration of the neural operator-based framework for generalized approximate Bayesian inference, showing the transformation from the prior distribution to the posterior distribution through data-driven training, conditioned on the observed data.}
  \label{fig:schematic}
\end{figure}

In this work, we leverage the advantages of neural operators and develop an equivariant neural operator framework for generalized approximate Bayesian inference, referred to as BI-EqNO, to address the limitations in both deterministic and stochastic methods. Through data-driven training, BI-EqNO transforms the prior distribution into the posterior distribution, conditioned on observed data, which is illustrated in Fig.~\ref{fig:schematic}. 
This transformation allows for flexible representations of both the prior and posterior distributions, as well as the observed data. 
Additionally, it preserves symmetry in the functional mapping by ensuring permutation equivariance between the prior and posterior representations and maintaining invariance to the ordering of observed data. 
Here, permutation equivariance means that if the input data is reordered, the model adjusts its predictions to maintain the correct correspondence between input and output values. Permutation invariance, on the other hand, ensures that reordering the observed data does not affect the model's predictions.
These properties are fundamental to Bayesian inference-based approaches and are essential for ensuring BI-EqNO's robustness across diverse applications. For example, in the structural health monitoring of a bridge, Gaussian processes can predict displacements throughout the structure using sensor data from various locations. If the prediction locations are reordered, the model updates the displacements accordingly, reflecting permutation equivariance. On the other hand, if the sequence of sensor measurements is rearranged, the predictions remain stable, demonstrating permutation invariance.

We demonstrate the flexibility and potential of BI-EqNO through the development of (1) the generalized Gaussian process (gGP) for regression and (2) the ensemble neural filter (EnNF) for sequential data assimilation. The generalized Gaussian process defines its mean and covariance functions using neural operators, effectively overcoming the limitations associated with predefined kernel structures in Gaussian process. Additionally, the ensemble neural filter functions as a generalized ensemble-based approach that can be trained and calibrated using data to enhance filtering performance, especially under conditions with limited ensemble sizes.

The remainder of this paper is organized as follows. Section~\ref{sec:method} introduces BI-EqNO framework and details its further development to the generalized Gaussian process for regression and the ensemble neural filter for sequential data assimilation. Section~\ref{sec:results} presents an evaluation of the generalized Gaussian process and the ensemble neural filter through representative case studies. Finally, Section~\ref{sec:conclude} offers a summary and concluding remarks.

\section{Methodology}
\label{sec:method}
\subsection{Description of BI-EqNO framework}
\label{sec:method-BI-EqNO}
\begin{figure}[!htb]
    \centering
    \includegraphics[width=0.85\textwidth]{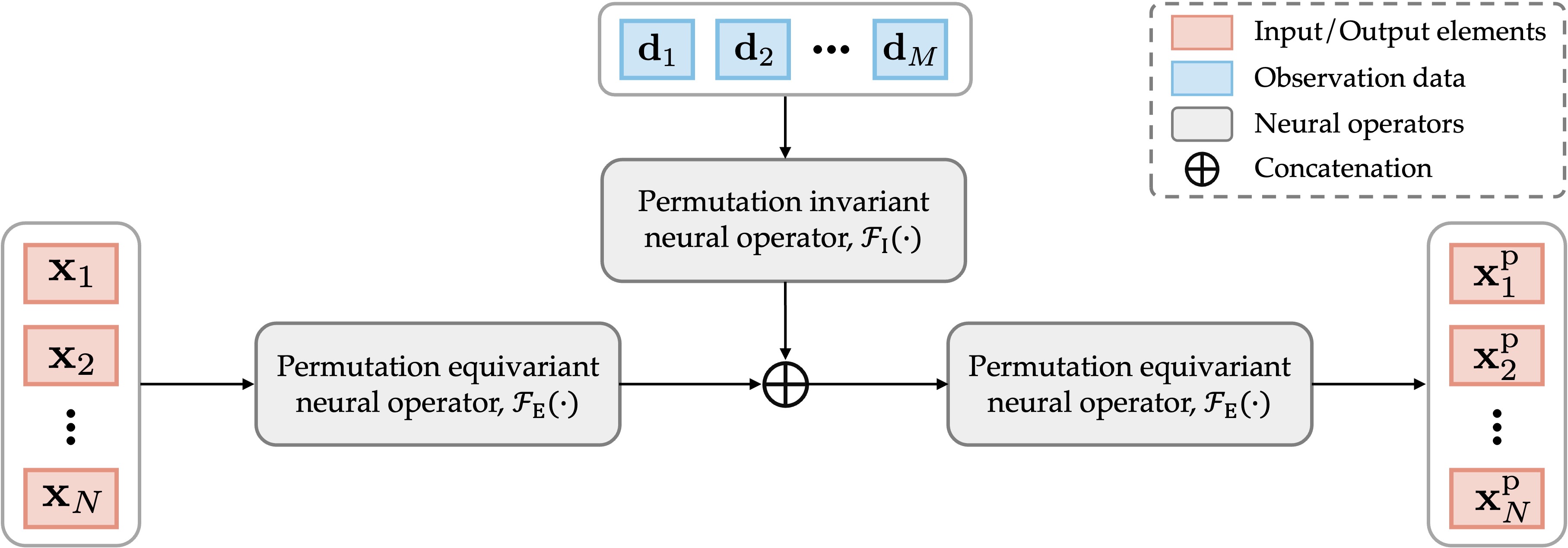}
    \caption{Schematic architecture of the equivariant neural operator framework for generalized approximate Bayesian inference. The input elements $\{\mathbf{x}_i\}_{i=1}^N$ are embedded through a permutation equivariant neural operator, while the observation data $\{\mathbf{d}_i\}_{i=1}^M$ are embedded through a permutation invariant neural operator. These embeddings are concatenated and passed through another permutation equivariant neural operator to generate the output $\{\mathbf{x}_i^\mathrm{P}\}_{i=1}^N$.}
    \label{fig:BI-EqNO}
\end{figure}

We propose the BI-EqNO framework to tackle challenges in approximate Bayesian inference. The schematic architecture of the BI-EqNO is illustrated in Fig.~\ref{fig:BI-EqNO}, showing the mapping from a collection of input elements $\{\mathbf{x}_i\}_{i=1}^N$ to the output $\{\mathbf{x}_i^\mathrm{P}\}_{i=1}^N$, conditioned on the observation data $\{\mathbf{d}_i\}_{i=1}^M$, through trainable neural operators. Before providing a detailed architectural description, we first present a high-level overview of its use across different scenarios.

\paragraph{Scenarios of BI-EqNO usage} 
The BI-EqNO framework is designed to transform a prior distribution to a posterior distribution conditioned on observations. However, it does not serve solely as a direct mapping between these two distributions; rather, its inputs and outputs are defined to suit different application scenarios. In the following, we present two representative use cases that illustrate its versatility.

\begin{enumerate}[(1)]
    \item \textbf{Prior-to-posterior mapping operator.} In this scenario, BI-EqNO functions as a mapping operator that directly transforms a prior distribution into a posterior distribution over the state space. Thus, the input and output of BI-EqNO are the corresponding samples or other representations of the state from the prior and posterior distributions.
    
    \item \textbf{Neural implicit representation of distributions.} Neural implicit representation refers to the use of neural networks to estimate a function continuously from its discrete representations. In this context, BI-EqNO maps coordinates (such as spatial locations, time points, or positions in feature space) into desired distribution representations (such as the mean and covariance of a Gaussian distribution). Consequently, the input to BI-EqNO consists of the coordinates, while the output corresponds to the representation of the prior or posterior distribution at these coordinates.
\end{enumerate}

\paragraph{BI-EqNO architecture} In either usage scenario, the BI-EqNO should maintain permutation equivariance between the input and output elements, while ensuring the numbering of the observations does not affect the output.
As shown in Fig.~\ref{fig:BI-EqNO}, to achieve this equivariance, the input elements $\{\mathbf{x}_i\}_{i=1}^N$ are embedded through a permutation equivariant neural operator $\mathcal{F}_{\text{E}}$ to preserve the ordering of the input elements. Simultaneously, the observations $\{\mathbf{d}_i\}_{i=1}^M$ are embedded through a permutation invariant neural operator $\mathcal{F}_{\text{I}}$, ensuring invariance to the ordering of the observations. The embeddings from both neural operators are then concatenated and passed through another permutation equivariant neural operator, serving as a fitting network, to produce the output $\{\mathbf{x}_i^\mathrm{P}\}_{i=1}^N$.

Mathematically, the permutation invariant neural operator, $\mathcal{F}_{\text{I}}: \left(\mathbb{R}^p\right)^N \rightarrow \mathbb{R}^q$, satisfies the property that for any reordering $\sigma$ of the indices \(\{1, 2, \ldots, N\}\) and elements $\mathbf{x}_1, \ldots, \mathbf{x}_N \in \mathbb{R}^p$, the following holds:
\begin{equation}
    \mathcal{F}_\text{I}(\mathbf{x}_{\sigma(1)}, \mathbf{x}_{\sigma(2)}, \ldots, \mathbf{x}_{\sigma(N)}) = \mathcal{F}_\text{I}(\mathbf{x}_1, \mathbf{x}_2, \ldots, \mathbf{x}_N).
    \label{eq:invariant}
\end{equation}

\begin{figure}[!htb]
    \centering
    \includegraphics[width=0.99\textwidth]{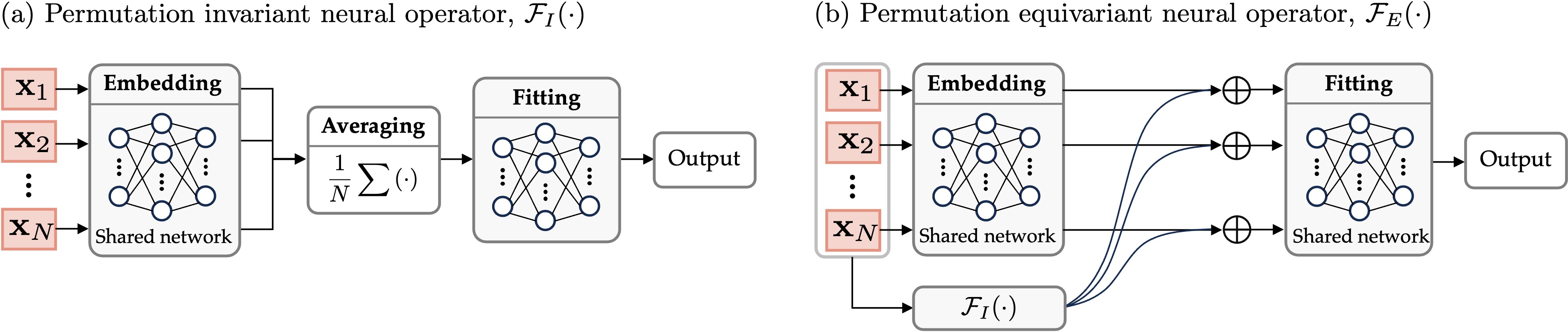}
    \caption{Architectures of (a) the permutation invariant neural operator $\mathcal{F}_\mathrm{I}$ and (b) the permutation equivariant neural operator $\mathcal{F}_\mathrm{E}$. In (a), each input element is individually processed through a shared embedding network. The resulting $N$ embeddings are then averaged to remove the influence of ordering, and the aggregated embedding is passed through a fitting network to produce the output. In (b), $\mathcal{F}_\mathrm{E}$ extends $\mathcal{F}_\mathrm{I}$ by concatenating element-wise embeddings to preserve the ordering of the input elements.}
    \label{fig:sub-operators}
\end{figure}

The architecture of the neural operator designed to achieve permutation invariance is illustrated in Fig.~\ref{fig:sub-operators}(a). This architecture employs an \textit{embedding-averaging-fitting} structure, which acts as a universal approximation of any permutation invariant function~\cite{han2019universal}. Specifically, the input elements $\{\mathbf{x}_i\}_{i=1}^N$ are firstly processed through a shared embedding network, which applies the same neural network to each element independently. These embeddings are then averaged to ensure the result is invariant to the ordering of the input elements. The averaged embedding is subsequently passed through a fitting network to produce the final output.

On the other hand, the permutation equivariant neural operator, $\mathcal{F}_{\text{E}}: \left( \mathbb{R}^p \right)^N \rightarrow \left( \mathbb{R}^{q} \right)^N$, satisfies the condition:
\begin{equation}
    \mathcal{F}_\text{E}(\mathbf{x}_{\sigma(1)}, \mathbf{x}_{\sigma(2)}, \ldots, \mathbf{x}_{\sigma(N)}) = \sigma \left( \mathcal{F}_\text{E}(\mathbf{x}_1, \mathbf{x}_2, \ldots, \mathbf{x}_N) \right).
    \label{eq:equivariant-eq}
\end{equation}

The designed neural operator that achieves this equivariance extends the architecture of the invariant operator by concatenating the element-wise embeddings through another shared embedding network, followed by a shared fitting network, as is illustrated in Fig.~\ref{fig:sub-operators}(b). The composite architecture, which combines a permutation invariant neural operator with a neural network acting on each input element, represents one type of approximation for permutation equivariant functions. However, it is not clear if this representation serves as a universal approximation.

\subsection{BI-EqNO as generalized Gaussian process}
\label{sec:method-gGP}

Gaussian processes are extensively used in various regression tasks to predict values at unobserved locations based on a limited set of observed data. However, their effectiveness is often constrained by the need for predefined covariance structures, which can limit their flexibility and adaptability to complex data patterns. To address this limitation, we employ the BI-EqNO as a neural implicit representation of prior and posterior distributions, providing a more flexible and trainable approach to covariance modeling. This innovation leads to the development of a generalized Gaussian process.

\paragraph{Regression} The goal of regression is to predict the values of one or more continuous target variables $y$ given a vector $\bm{x}$ of input variables. Given a dataset $\mathcal{D} = \{X, Y\}$ of $M$ observations with input values
$X = \left[\bm{x}^\text{dat}_1, \bm{x}^\text{dat}_2, \ldots, \bm{x}^\text{dat}_M\right]$ and corresponding target values $Y = \left[\tilde{y}^\text{dat}_1, \tilde{y}^\text{dat}_2, \ldots, \tilde{y}^\text{dat}_M\right]^\top$, the task is to predict the target values $Y^*$ for new input values $X^* = \left[\bm{x}^*_1, \bm{x}^*_2, \ldots, \bm{x}^*_N\right]$, where $N$ is the number of test points. Here, the target variable $y$ is assumed to be a scalar.

\paragraph{Regression with Gaussian processes} Gaussian processes provide a probabilistic framework for regression by defining a distribution over the underlying functions $f$. The key assumption is that the function values at any finite set of inputs are jointly Gaussian. Specifically, the steps in Gaussian process regression are as follows:
\begin{enumerate}[(1)]
    \item Specify a prior distribution as a Gaussian process:
    \begin{equation}
    f(\bm{x}) \sim \mathcal{GP} \left(\mathsf{m}(\bm{x}), \mathsf{K}(\bm{x}, \bm{x}')\right),
    \end{equation}
    where $\mathsf{m}(\bm{x})$ denotes the mean function and $\mathsf{K}(\bm{x}, \bm{x}')$ denotes the kernel (or covariance) function, which together characterize the Gaussian distribution.
    
    \item Formulate the joint distribution between the observed target values $Y$ and the function values $Y^* = f(X^*)$ at test points under the defined prior:
    \begin{equation}
    \begin{bmatrix}
    Y \\
    Y^*
    \end{bmatrix}
    \sim \mathcal{N} \left( 
    \begin{bmatrix}
    \mathsf{m}(X)\\
    \mathsf{m}(X^*)
    \end{bmatrix}, 
    \begin{bmatrix}
    \mathsf{K}(X, X) + \sigma_{\varepsilon}^2 \mathbf{I} & \mathsf{K}(X, X^*) \\
    \mathsf{K}(X^*, X) & \mathsf{K}(X^*, X^*)
    \end{bmatrix}
    \right),
    \label{eq:joint-distri}
    \end{equation}
    where the observed target values are assumed to be corrupted by Gaussian noise $\epsilon$ with variance $\sigma_\epsilon^2$, i.e., $\tilde{y}_i^\text{dat} = f(\bm{x}_i^\text{dat}) + \epsilon_i$, with $\epsilon_i \sim \mathcal{N}(0, \sigma_{\epsilon}^2)$.

    \item Compute the posterior distribution of $Y^*$, conditioned on the training dataset $\cD$, using Bayes's theorem:
    \begin{equation}
    p\left(Y^* | X^*, \cD\right)=\frac{p\left(Y^*, Y | X, X^*\right)} {p\left(Y | X \right)}.
    \label{eq:Bayes-GPR}
    \end{equation}
    This results in the conditional posterior Gaussian distribution $p\left(Y^* | X^*, \cD\right)=\mathcal{N}\left(\mathsf{m}_\cD(X^*), \mathsf{K}_\mathcal{D}(X^*, X^*)\right)$, with mean $\mathsf{m}_\mathcal{D}(X^*)$ and covariance $\mathsf{K}_\mathcal{D}(X^*, X^*)$ as follows:
    \begin{equation}
    \begin{aligned}
    \mathsf{m}_\mathcal{D}(X^*) & = \mathsf{m}(X^*) + \mathsf{K}(X^*, X)\left[\mathsf{K}(X, X)+\sigma_\epsilon^2 \mathbf{I}\right]^{-1} \left[Y - \mathsf{m}(X)\right], \\
    \mathsf{K}_\mathcal{D}(X^*, X^*) & =\mathsf{K}(X^*, X^*)-\mathsf{K}(X^*, X)\left[\mathsf{K}(X, X)+\sigma_\epsilon^2 \mathbf{I}\right]^{-1} \mathsf{K}(X, X^*).
    \end{aligned}
    \label{eq:GPR-post}
    \end{equation}
\end{enumerate}
From Eq.~\eqref{eq:GPR-post}, it is evident that the posterior depends on the kernel function $\mathsf{K}$: Both the kernel functional form and the associated hyperparameters determine the model's efficacy. The kernel functional form, such as the squared exponential kernel, is usually predetermined and remains fixed throughout the inference process, while the hyperparameters $\bm{\varphi}$ that control the kernel's length scale and variance can be optimized by maximizing the log marginal likelihood, i.e., 
\begin{equation}
    \bm{\varphi}_\text{opt} = \underset{\bm{\varphi}}{\operatorname{argmax}} \log p(Y | X, \bm{\varphi}).
    \label{eq:GP-optimize}
\end{equation}
However, selecting an appropriate kernel can be particularly challenging in real-world scenarios that involve discontinuities, such as sudden changes in signal, or multiple spatial scales, like varying levels of detail in geospatial data.

\paragraph{Generalized Gaussian process: Requirements} To address the difficulty in kernel selection, we develop the generalized Gaussian process based on the BI-EqNO framework. Unlike the Gaussian process that relies on predefined kernels, the generalized Gaussian process uses a neural operator-based mean function $\mathsf{m}^\text{NN}(\bm{x})$ and covariance function $\mathsf{K}^\text{NN}(\bm{x}, \bm{x}')$ to define the function distribution, i.e., $f(\bm{x}) \sim \mathcal{GP} \left(\mathsf{m}^\text{NN}(\bm{x}), \mathsf{K}^\text{NN}(\bm{x}, \bm{x}')\right)$, allowing for greater flexibility and adaptability. Note that directly applying BI-EqNO to represent the generalized Gaussian process poses challenges due to the strict requirements for the outputs---the mean vector $\mathsf{m}^\text{NN}(X^*)$ and the covariance matrix $\mathsf{K}^\text{NN}(X^*, X^*)$. These include maintaining proper dimensionality of matrices, ensuring permutation invariance and equivariance with respect to observation and test points, respectively, and guaranteeing symmetry and positive semi-definiteness of the covariance matrix. Specifically, they must satisfy the following mathematical constraints:
\begin{enumerate}[(1)]
    \item Their dimensions should adapt to the number of test points in $X^*$. Specifically, $\mathsf{m}^\text{NN}(X^*) \in \mathbb{R}^N$ and $\mathsf{K}^\text{NN}(X^*, X^*) \in \mathbb{R}^{N \times N}$.
    \item Both $\mathsf{m}^\text{NN}(X^*)$ and $\mathsf{K}^\text{NN}(X^*, X^*)$ must be permutation invariant to the observation points $X$ and permutation equivariant to the test points $X^*$, as implied by Eq.~\eqref{eq:GPR-post} and detailed in~\ref{app:gGP-constraints}.
    \item The covariance matrix $\mathsf{K}^\text{NN}(X^*, X^*)$ should be symmetric and positive semi-definite.
\end{enumerate}
To satisfy these constraints, we represent the output of covariance as:
\begin{equation}
\mathsf{K}^\text{NN}(X^*, X^*) = LL^{\top} + \exp(D),
\label{eq:cov-represent}
\end{equation}
where $L = [\bm{l}_1, \ldots, \bm{l}_N]^\top$ is a low-rank matrix and $D = \operatorname{diag}\left(d_1, \ldots, d_N\right)$ is a diagonal matrix. The latent vectors $\{\bm{l}_i\}_{i=1}^N$ and the diagonal elements $\{d_i\}_{i=1}^N$ must adhere to the second constraint as well.

\begin{figure}[!htb]
    \centering
    \includegraphics[width=0.99\textwidth]{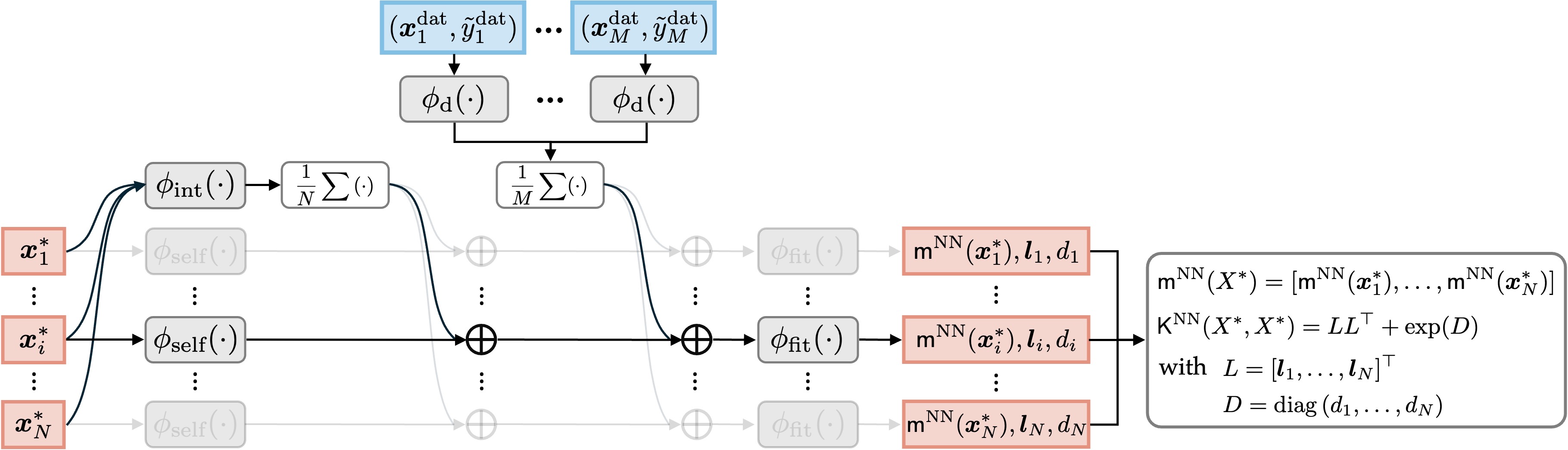}
    \caption{Architecture of the generalized Gaussian process (gGP) for regression. It defines a permutation equivariant mapping from the test points $X^* = \{\bm{x}_i^*\}_{i=1}^N$ to the mean $\mathsf{m}^\mathrm{NN}(X^*)$ and covariance $\mathsf{K}^\mathrm{NN}(X^*, X^*)$ of their function values, conditioned on the observation data $\mathcal{D} = \{(\bm{x}_i^\text{dat},\tilde{y}_i^\text{dat})\}_{i=1}^{M}$ in a permutation invariant manner. 
    From left to right: The test points $X^*$ are processed through two shared neural networks, $\phi_\text{int}$ for interaction embedding and $\phi_\text{self}$ for self-embedding. Meanwhile, the observation data $\mathcal{D}$ is embedded through a share network $\phi_\mathrm{d}$. The embeddings from both test points and observation data are concatenated and passed through a shared fitting network $\phi_\text{fit}$ to generate the predicted function values $\{\mathsf{m}^\text{NN}(\bm{x}_i^*)\}_{i=1}^N$ and the covariance matrix $\mathsf{K}^\text{NN}(X^*, X^*)$, based on the latent vectors $\{\bm{l}_i\}_{i=1}^N$ and the diagonal elements $\{d_i\}_{i=1}^N$.}
    \label{fig:gGP}
\end{figure}

\paragraph{Generalized Gaussian process: Network architecture} 
In view of the requirements, the architecture is designed to balance structural simplicity with functional flexibility while imposing the additional constraints. The architecture incorporates both a permutation equivariant neural operator for the test points and a permutation invariant neural operator for the observation data. Specifically, as shown in Fig.~\ref{fig:gGP}, the test points $\{\bm{x}_i^*\}_{i=1}^{N}$ are embedded through a permutation equivariant neural operator, which is built upon two shared neural networks: $\phi_\text{int}$ for interaction embedding and $\phi_\text{self}$ for self-embedding. The observation data $\{(\bm{x}_i^\text{dat},\tilde{y}_i^\text{dat})\}_{i=1}^{M}$ are embedded using a permutation invariant neural operator, based on a shared network $\phi_\text{d}$. To simplify the architecture, the design omits the fitting components from the individual operators, as introduced earlier in Fig.~\ref{fig:sub-operators}. Instead, the embeddings from both operators are concatenated and passed through a shared fitting neural network $\phi_\text{fit}$. This fitting network generates the final outputs: the mean values $\{\mathsf{m}^\text{NN}(\bm{x}_i^*)\}_{i=1}^N$, the latent vectors $\{\bm{l}_i\}_{i=1}^N$ and the diagonal elements $\{d_i\}_{i=1}^N$. Mathematically, the output is given by:
\begin{equation}
    \{\mathsf{m}^\text{NN}(\bm{x}_i^*), \bm{l}_i, d_i\} = \phi_\text{fit}\left(
    \phi_\text{self}\left(\bm{x}_i^*\right) \oplus \frac{1}{N} \sum_{j=1}^N \phi_\text{int}\left(\bm{x}_j^*\right) \oplus \frac{1}{M} \sum_{k=1}^M \phi_\text{d}\left(\bm{x}_k^\text{dat}, \tilde{y}_k^\text{dat}\right)
    \right) \quad \text{for} \quad i \in\{1, \ldots, N\}.
\end{equation}

\paragraph{Generalized Gaussian process: Training} The generalized Gaussian process is trained by requiring it to predict a randomly selected subset of $\mathcal{D}$, denoted as $\mathcal{D}' = \{X', Y'\}$, conditioned on the remaining set $\cD \backslash \cD'$ as observations. The subset size can be flexible but fixed in this work for training efficiency. The parameters associated with the generalized Gaussian process, specifically the weights $\bm{w}$ in $\phi_\text{self}$, $\phi_\text{int}$, $\phi_\text{d}$, and $\phi_\text{fit}$, are optimized by minimizing the negative conditional log probability:
\begin{equation}
\begin{aligned}
\bm{w}_\text{opt} &= \underset{\bm{w}}{\operatorname{argmin}} \; \left[\frac{1}{2}\left(Y' - \mathsf{m}^\text{NN}(X';\bm{w})\right)^\top {\mathsf{K}^\text{NN}(X', X';\bm{w})}^{-1} \left(Y' - \mathsf{m}^\text{NN}(X';\bm{w})\right) \right.\\
&\left.\qquad \qquad \quad + \frac{1}{2} \log |\mathsf{K}^\text{NN}(X', X';\bm{w})| + \frac{M'}{2} \log (2 \pi)\right],
\end{aligned}
\label{eq:gGP-optimize}
\end{equation}
where $|\cdot|$ denotes the determinant of a matrix and $M'$ is the number of points in $X'$. After training, the posterior distribution of the function values at test points $X^*$ becomes
\begin{equation}
p\left(Y^* | X^*, \cD\right)=\mathcal{N}\left(\mathsf{m}^\text{NN}(X^*), \mathsf{K}^\text{NN}(X^*, X^*); \bm{w}_\text{opt}\right),
\end{equation}
where the mean vector and covariance matrix are fully determined by the optimized weights $\bm{w}_\text{opt}$, given the entire set of the observations $\cD$.

\paragraph{Comparison to Gaussian process}
We compare the Gaussian process (GP) with the generalized Gaussian process (gGP) for regression, as summarized in Table~\ref{table:compare-GP-gGP}.
In the standard GP framework, the prior distribution of $f(X^*)$ is characterized by the mean vector $\mathsf{m}(X^*)$ and the covariance matrix $\mathsf{K}(X^*, X^*; \bm{\varphi}_\text{init})$, where $\bm{\varphi}_\text{init}$ denotes the initial hyperparameters of a predefined kernel function. The posterior distribution is analytically derived via Bayes' theorem, updating the mean and covariance to $\mathsf{m}_\cD(X^*)$ and $\mathsf{K}_\cD(X^*, X^*)$, respectively, which still depend on the initial hyperparameters $\bm{\varphi}_\text{init}$. Further optimization adjusts these hyperparameters to $\bm{\varphi}_\text{opt}$.

In contrast, gGP employs neural operators to model the mean and covariance, denoted as $\mathsf{m}^\text{NN}(X^*)$ and $\mathsf{K}^\text{NN}(X^*, X^*)$, both parameterized by the neural operator weights $\bm{w}$. The posterior distribution is obtained via data-driven training, with the weights being optimized from initial values $\bm{w}_\text{init}$ to $\bm{w}_\text{opt}$. This comparison underscores key methodological differences and optimization approaches: GP relies on hyperparameter tuning with a predefined kernel structure, while gGP leverages neural networks for enhanced flexibility in kernel representation.

\begin{table}[ht]
\centering
\renewcommand{\arraystretch}{1.5}
\small 
\begin{tabular}{|p{1.5cm}|c|c|c|}
\hline
\multicolumn{2}{|c|}{} & \textbf{Gaussian process (GP)} & \textbf{Generalized Gaussian process (gGP)} \\ \hline
\multirow{3}{=}[-2.0em]{\centering \textbf{Bayesian inference}} & \begin{tabular}[c]{@{}c@{}} \textbf{Prior} \\ $p(f(X^*) | X^*)$ \end{tabular} & $\mathcal{N}\left(\mathsf{m}(X^*), \mathsf{K}(X^*, X^*; \bm{\varphi}_\text{init})\right)$ & $\mathcal{N}\left(\mathsf{m}^\text{NN}(X^*), \mathsf{K}^\text{NN}(X^*, X^*); \bm{w}_\text{init}\right)$ \\ \cline{2-4} 
 & \begin{tabular}[c]{@{}c@{}} \textbf{Posterior} \\ $p(f(X^*) | X^*, \mathcal{D})$ \end{tabular} & \begin{tabular}[c]{@{}c@{}} $\mathcal{N}\left(\mathsf{m}_\cD(X^*), \mathsf{K}_\cD(X^*, X^*); \bm{\varphi}_\text{init}\right)$ \\ obtained by Bayes' theorem, see Eq.~\eqref{eq:GPR-post} \end{tabular} & \begin{tabular}[c]{@{}c@{}} $\mathcal{N}\left(\mathsf{m}^\text{NN}(X^*), \mathsf{K}^\text{NN}(X^*, X^*); \bm{w}_\text{opt}\right)$ \\ obtained by training neural networks \end{tabular} \\ \cline{2-4}
 & \begin{tabular}[c]{@{}c@{}} \makecell{\textbf{Posterior} \\ \textbf{with optimization}}
\end{tabular} & \begin{tabular}[c]{@{}c@{}} $\mathcal{N}\left(\mathsf{m}_\cD(X^*), \mathsf{K}_\cD(X^*, X^*); \bm{\varphi}_\text{opt}\right)$ \end{tabular} & \begin{tabular}[c]{@{}c@{}} --- \\ (same as the above cell) \end{tabular} \\ \hline
\multicolumn{2}{|c|}{\textbf{Optimization}} & \begin{tabular}[c]{@{}c@{}}Hyperparameters $\bm{\varphi}$ with\\ predefined kernel functional form \end{tabular} & \begin{tabular}[c]{@{}c@{}}Neural network weights $\bm{w}$ with \\ flexible kernel functional form \end{tabular} \\ \hline
\end{tabular}
\caption{Comparison of Gaussian process (GP) and generalized Gaussian process (gGP) for regression. The comparison highlights the methodological differences, focusing on the flexibility of neural networks in gGP, which allows for a more adaptable and flexible kernel representation compared to the predefined kernel structure in GP.}
\label{table:compare-GP-gGP}
\end{table}

\subsection{BI-EqNO as ensemble neural filter}
\label{sec:method-EnNF}
We employ the BI-EqNO as a surrogate operator to update the state of a dynamic system with new observation data, leading to the ensemble neural filter (EnNF). This method provides a flexible, nonlinear update scheme that can potentially represent various ensemble-based filters, such as the ensemble Kalman filter (EnKF), and demonstrates certain advantages in small-ensemble scenarios.

\paragraph{Sequential data assimilation}
Sequential data assimilation involves continuously updating the state of a dynamic system as new observations become available. The goal is to combine model predictions and observational data to improve the state estimate over time. The state evolution is typically represented by a dynamic model, $\mathbf{z}(t_k)=\mathcal{M}\left(\mathbf{z}(t_{k-1})\right)+\mathbf{w}(t_k)$, where $\mathbf{z}(t_k)$ is the system's state at time $t_k$, $\mathcal{M}$ is the model operator, and $\mathbf{w}(t_k)$ is the model error. Observations are related to the state through an observation model, $\mathbf{d}(t_k)=\mathcal{H}\left(\mathbf{z}(t_k)\right)+\mathbf{v}(t_k)$, where $\mathbf{d}(t_k)$ is the observation vector, $\mathcal{H}$ is the observation operator mapping the state vector into observation space, and $\mathbf{v}(t_k)$ is the observation error. At each time step, the model's forecast of the state is updated using the latest observations, incorporating uncertainties in both the model and the observations. This iterative process refines the state estimate to ensure it remains as accurate as possible with new data.

\paragraph{Data assimilation with ensemble Kalman filter}
The ensemble Kalman filter (EnKF) is a sequential Monte Carlo method designed for sequential data assimilation in complex, high-dimensional systems, such as those found in geosciences and hydrology. It leverages a finite ensemble of states, comprising significantly fewer realizations than the state dimension, to approximate state uncertainty. Specifically, the EnKF maintains an ensemble of state realizations, $\mathbf{Z}=\left(\mathbf{z}_1, \mathbf{z}_2, \ldots, \mathbf{z}_N\right) \in \mathbb{R}^{n \times N}$, where each realization represents a possible state of the system. During an assimilation window, each realization is propagated through a numerical model to generate the forecast ensemble, $\mathbf{Z}^{\mathrm{f}}=\left(\mathbf{z}_1^{\mathrm{f}}, \mathbf{z}_2^{\mathrm{f}}, \ldots, \mathbf{z}_N^{\mathrm{f}}\right)$. These forecasts, also known as the prior ensemble, are then updated based on observations, $\mathbf{D}=\left(\mathbf{d}_1, \mathbf{d}_2, \ldots, \mathbf{d}_N\right) \in \mathbb{R}^{m \times N}$, to obtain the posterior ensemble, $\mathbf{Z}^{\mathrm{a}}=\left(\mathbf{z}_1^{\mathrm{a}}, \mathbf{z}_2^{\mathrm{a}}, \ldots, \mathbf{z}_N^{\mathrm{a}}\right)$. The update equation for each realization follows:
\begin{equation}
\mathbf{z}_i^{\mathrm{a}}=\mathbf{z}_i^{\mathrm{f}}+ \mathbf{K}\left(\mathbf{d}_i-\mathcal{H}(\mathbf{z}_i^{\mathrm{f}}\right)) \quad \text{with} \quad \mathbf{K} = \overline{\mathbf{C}}_{z y}\left(\overline{\mathbf{C}}_{y y} + \overline{\mathbf{C}}_{d d}\right)^{-1},
\label{eq:EnKF-update}
\end{equation}
where $\mathbf{K}$ is the Kalman gain matrix, $\overline{\mathbf{C}}_{z y}$ is the ensemble covariance matrix between the state vector $\mathbf{z}$ and the predicted observation $\mathbf{y} = \mathcal{H}(\mathbf{z})$, $\overline{\mathbf{C}}_{y y}$ is the ensemble covariance matrix of predicted observation, and $\overline{\mathbf{C}}_{d d}$ is the ensemble covariance matrix of observation errors.

As suggested by Eq.~\eqref{eq:EnKF-update}, the assimilation performance of EnKF relies on accurate covariance estimation, which necessitates a sufficiently large ensemble size $N$. In practical applications, however, computational limitations often restrict the ensemble size, leading to issues such as sampling errors, underestimation of state variability, and spurious correlations. To address these challenges, techniques such as localization are employed, which limit the influence of distant observations on the state updates. While effective, these techniques can be complex to implement and must be tailored to specific problems.

\paragraph{Ensemble neural filter: Requirements}
Our goal is to develop a ``super'' filter capable of representing a wide range of ensemble-based filters, while being easily calibrated through data-driven training. A straightforward design approach involves using the prior and posterior state ensembles as the input and output of the BI-EqNO. However, this approach, which entails a many-to-many full-field mapping, can result in high memory demands in high-dimensional systems and pose challenges during the training process. To overcome these issues, we redesign the filter, drawing inspiration from the core principles of the ensemble Kalman filter. The redesigned filter, referred to as the ensemble neural filter, adheres to the following requirements:
\begin{enumerate}[(1)]
    \item The ensemble neural filter is flexible with arbitrary ensemble sizes.
    \item The permutations of prior and posterior realizations are equivariant.
    \item The update of each state variable shares the same ensemble neural filter.
    \item The ensemble neural filter depends only on the information within the observation space, consisting of observations, $\{\mathbf{d}_i\}_{i=1}^{N}$, and predicted observations, $\{\mathcal{H}(\mathbf{z}_i^\text{f})\}_{i=1}^{N}$.
\end{enumerate}
The first two requirements are straightforward to understand and implement, while the latter two are more complex and are further detailed in~\ref{app:EnNF-properties}.

\begin{figure}[!htb]
\centering
\includegraphics[width=0.75\textwidth]{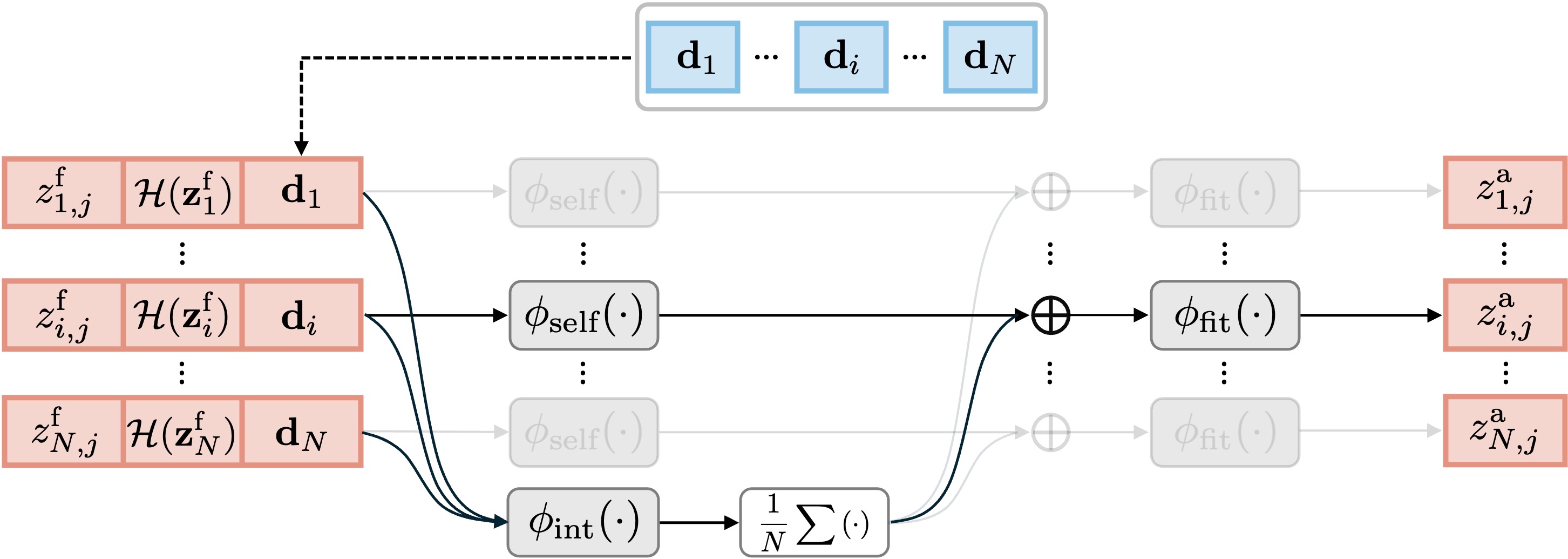}
  \caption{Architecture of the ensemble neural filter (EnNF) for sequential data assimilation. The EnNF defines a permutation equivariant mapping from prior to posterior ensembles of individual state variables, i.e., $\{z_{i,j}^\mathrm{f}\}_{i=1}^N \mapsto \{z_{i,j}^\mathrm{a}\}_{i=1}^N$, uniquely conditioned on the information within the observation space, $\left\{\left(\mathcal{H} (\mathbf{z}_i^\text{f}), \mathbf{d}_i\right)\right\}_{i=1}^N$. The ensemble statistical properties and the realization-specific properties are embedded through a shared network $\phi_\mathrm{int}$ and another shared network $\phi_\mathrm{self}$, respectively. These embeddings are concatenated and passed through a shared fitting network $\phi_\mathrm{fit}$ to yield the updated state variables.}
  \label{fig:EnNF}
\end{figure}

\paragraph{Ensemble neural filter: Network architecture} In light of the requirements, the architecture of the ensemble neural filter (EnNF) is designed to map between the prior and posterior ensembles of individual state variables for memory efficiency, while ensuring the fulfillment of other constraints. Specifically, the EnNF maps between $\left\{z_{i,j}^{\text{f}}\right\}_{i=1}^N$ and $\left\{z_{i,j}^{\text{a}}\right\}_{i=1}^N$, representing the prior and posterior ensembles of the $j$-th state variable, with this mapping shared across all $j \in \{1, \ldots, n\}$. Moreover, this mapping is uniquely conditioned on the information within the observation space. To achieve this, both the observations $\{\mathbf{d}_i\}_{i=1}^{N}$ and the predicted observations $\{\mathcal{H}(\mathbf{z}_i^\text{f})\}_{i=1}^{N}$ are integrated into the input of EnNF. The integration process preserves permutation equivariance, as the update of each realization depends on both the statistical properties of the integrated data and the specific data corresponding to each realization. As a result, the EnNF defines a permutation equivariant mapping from $\left\{\left(z_{i,j}^{\text{f}}, \mathcal{H} (\mathbf{z}_i^\text{f}), \mathbf{d}_i\right)\right\}_{i=1}^N$ to $\left\{z_{i,j}^{\text{a}}\right\}_{i=1}^N$ for all $j \in \{1, \ldots, n\}$, which can be achieved using a single permutation equivariant neural operator, as shown in Fig.~\ref{fig:EnNF}. The structure directly follows the design in Fig.~\ref{fig:sub-operators}(b), and therefore does not require further detailed description.
Mathematically, the EnNF update equation is written as:
\begin{equation}
    z_{i,j}^{\text{a}} = \phi_{\text{fit}} \left( \phi_{\text{self}} \left( z_{i,j}^{\text{f}}, \mathcal{H}(\mathbf{z}_i^{\text{f}}), \mathbf{d}_i \right) \oplus \frac{1}{N} \sum_{i=1}^N \phi_{\text{int}} \left( z_{i,j}^{\text{f}}, \mathcal{H}(\mathbf{z}_i^{\text{f}}), \mathbf{d}_i \right) \right) \quad \text{for} \quad i \in\{1, \ldots, N\} \quad \text{and} \quad j \in\{1, \ldots, n\}.
    \label{eq:EnNF-update}
\end{equation}

\paragraph{Comparison to EnKF}
We compare the ensemble Kalman filter (EnKF) and the ensemble neural filter (EnNF) for sequential data assimilation, as summarized in Table~\ref{table:compare-EnKF-EnNF}. Here, both filtering methods assume Gaussian distributions for the prior and posterior distributions of the state vector $\mathbf{z}$, though EnNF does not inherently require this assumption and can be adapted for non-Gaussian scenarios. The mean is obtained by averaging the ensemble members, while the covariance is computed from the spread of the ensemble members around the mean. The key difference lies in the update formula. The EnKF update equation modifies each realization $\mathbf{z}_i^\text{f}$ using the Kalman gain $\mathbf{K}$ and the innovation (the difference between the observation $\mathbf{d}_i$ and predicted observation $\mathcal{H}(\mathbf{z}_i^\text{f})$). This leads to a linear update scheme (see \ref{app:EnNF-properties}), where the updated realizations are confined to the subspace spanned by the prior ones.
In contrast, the EnNF employs an equivariant neural operator to update the state, offering a more general filtering scheme that can flexibly adapt to data. Due to its nonlinearity, the EnNF allows for updates that can step outside the subspace spanned by the prior ones, potentially providing more accurate state updates in complex scenarios.

\begin{table}[ht]
\centering
\renewcommand{\arraystretch}{1.5}
\small 
\begin{tabular}{|p{1.5cm}|c|c|c|}
\hline
\multicolumn{2}{|c|}{} & \textbf{Ensemble Kalman filter (EnKF)} & \textbf{Ensemble neural filter (EnNF)} \\ \hline
\multirow{2}{=}[-1.2em]{\centering \textbf{Bayesian inference}} & \begin{tabular}[c]{@{}c@{}} \textbf{Prior} \\ $p(\mathbf{z})$ \end{tabular} & \multicolumn{2}{c|}{\begin{tabular}[c]{@{}c@{}} $\mathcal{N}\left(\overline{\mathbf{z}}^\text{f}, \mathbf{P}^\text{f}\right)$, \\ $\text{with }\, \overline{\mathbf{z}}^\text{f}=\frac{1}{N} \sum_{i=1}^N \mathbf{z}_i^\text{f} \, \text{ and } \, \mathbf{P}^\text{f}=\frac{1}{N-1} \sum_{i=1}^N\left(\mathbf{z}_i^\text{f}-\overline{\mathbf{z}}^\text{f}\right)\left(\mathbf{z}_i^\text{f}-\overline{\mathbf{z}}^\text{f}\right)^\top$ \end{tabular}} \\ \cline{2-4} 
& \begin{tabular}[c]{@{}c@{}} \textbf{Posterior} \\ $p(\mathbf{z} | \mathcal{D})$ \end{tabular} & \multicolumn{2}{c|}{\begin{tabular}[c]{@{}c@{}} $\mathcal{N}\left(\overline{\mathbf{z}}^\text{a}, \mathbf{P}^\text{a}\right)$, \\ $\text{with }\, \overline{\mathbf{z}}^\text{a}=\frac{1}{N} \sum_{i=1}^N \mathbf{z}_i^\text{a} \, \text{ and } \, \mathbf{P}^\text{f}=\frac{1}{N-1} \sum_{i=1}^N\left(\mathbf{z}_i^\text{a}-\overline{\mathbf{z}}^\text{a}\right)\left(\mathbf{z}_i^\text{a}-\overline{\mathbf{z}}^\text{a}\right)^\top$ \end{tabular}} \\ \hline
\multicolumn{2}{|c|}{\textbf{Analysis}} & \begin{tabular}[c]{@{}c@{}}$\mathbf{z}_i^\text{a}=\mathbf{z}_i^\text{f}+ \mathbf{K}\left(\mathbf{d}_i-\mathcal{H}(\mathbf{z}_i^\text{f}\right))$ \end{tabular}& \begin{tabular}[c]{@{}c@{}}$\left\{z_{i, j}^{\text{a}}\right\}_{i=1}^N=\mathcal{F}_\text{E}\left(\left\{\left(z_{i, j}^{\text{f}}, \mathcal{H}(\mathbf{z}_i^\text{f}), \mathbf{d}_i\right)\right\}_{i=1}^N\right)$\end{tabular} \\ \hline
\end{tabular}
\caption{Comparison of ensemble Kalman filter (EnKF) and ensemble neural filter (EnNF) for sequential data assimilation. Here, both methods assume Gaussian distributions for the prior and posterior of the state vector, with the mean and covariance computed from the ensemble members. EnNF highlights its flexibility and adaptability to data by using a trainable neural operator for analysis.}
\label{table:compare-EnKF-EnNF}
\end{table}

\section{Results}
\label{sec:results}
\subsection{Generalized Gaussian process (gGP) for regression}
We numerically investigate the regression performance of the generalized Gaussian process for learning two representative functions: a one-dimensional discontinuous function and a two-dimensional linear combination of truncated trigonometric functions. The discontinuous function is characterized by its non-smoothness with abrupt changes, while the latter function exhibits a multi-scale property. These features pose significant challenges in selecting appropriate kernel functions when using Gaussian processes. In contrast, the kernel functions in generalized Gaussian process can adapt flexibly to observational data through data-driven training, leading to superior regression performance compared to Gaussian processes.

\subsubsection{Regression on a one-dimensional discontinuous function}
\label{sec:results-gGP-stepfunc}

\paragraph{Target function}
Discontinuous functions present a challenging test case for regression models due to their abrupt changes and discontinuities. Gaussian processes often fail to accurately capture these features because of the smoothness assumptions inherent in commonly used kernels, such as the squared exponential kernel. As a specific example, the one-dimensional discontinuous function (i.e., one-dimensional indicator function) employed in this study is defined as follows:
\begin{equation}
f(x)= \begin{cases}0, & \text { if } 0 \leq x<0.3, \\ 1, & \text { if } 0.3 \leq x<0.7, \\ 0, & \text { if } 0.7 \leq x \leq 1.\end{cases}
\label{eq:1D-stepfunc}
\end{equation}

\paragraph{Experiment setup} 
The objective is to predict the function values at 1,000 uniformly distributed test points within the interval $[0,1]$ using the provided observation data. The observation data $\mathcal{D}$ comprises input values $X = \left[x^\text{dat}_1, \ldots, x^\text{dat}_M\right]$ uniformly sampled from this interval, with corresponding target values $Y = \left[\tilde{y}^\text{dat}_1, \ldots, \tilde{y}^\text{dat}_M\right]^\top$ assumed to be corrupted by Gaussian noise:
\begin{equation}
    \tilde{y}_i^\text{dat} = f(x_i^\text{dat}) + \epsilon_i,
\end{equation}
where $\epsilon_i \sim \mathcal{N}(0, 0.01^2)$. We consider two observation conditions: a sparse scenario with 30 observation points ($M = 30$) and a dense scenario with 100 observation points ($M=100$).

For each observation condition, both Gaussian process and generalized Gaussian process are employed to perform the regression task. In the Gaussian process, a squared exponential kernel is utilized. Specifically, it is defined as $\mathsf{K}(x, x') = \sigma_f^2 \exp\left[-\frac{1}{2} (x - x')^2/\ell^2 \right]$, where the hyperparameters, including the length scales $\ell$ and the variance $\sigma_f$, are optimized by maximizing the log marginal likelihood. In the generalized Gaussian process, the neural operator is trained on the observation data, which involves optimizing the same likelihood-based criterion to ensure accurate function approximation.

\paragraph{Data generation} To train the generalized Gaussian process, we define the task as predicting a randomly selected subset $\cD' = \{X', Y'\}$ from the entire dataset $\cD$, conditioned on the remaining set $\cD \backslash \cD'$. Specifically, the training data consists of input-output pairs, where the input includes the test points $X'$ and the observations from $\cD \backslash \cD'$, and the output corresponds to the observed target values $Y'$. The set $X'$ contains $M'(< M)$ test points. Under the sparse observation condition, $M'$ is set to 20, with the remaining 10 points used as observed points. Under the dense observation condition, $M'$ is set to 10, with the remaining 90 points used as observed points. For both observation conditions, we randomly sample $2 \times 10^4$ data pairs without replacement, ensuring a comprehensive dataset for model training. With the training dataset, we train the generalized Gaussian process by minimizing the negative conditional log probability, as shown in Eq.~\eqref{eq:gGP-optimize}.

\begin{figure}[!htb]
    \centering
    \includegraphics[width=0.99\linewidth]{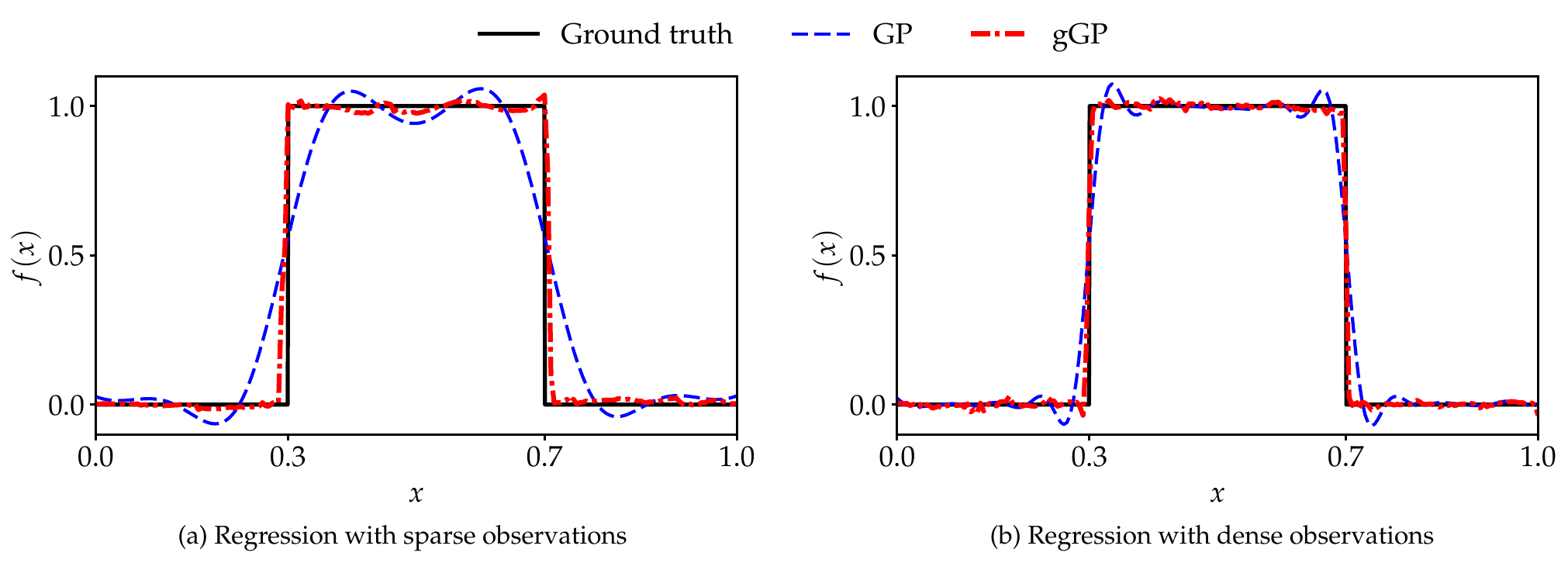}
    \caption{Comparison of regression performance for the one-dimensional discontinuous function using the Gaussian process (GP) and generalized Gaussian process (gGP) with sparse and dense observations. Panel (a) shows predictions with 30 observations, while panel (b) shows predictions with 100 observations. The GP produces smooth transitions in the discontinuous regions, whereas the gGP accurately captures the abrupt changes.}
    \label{fig:stepfunc-mean-covariance}
\end{figure}

\paragraph{Regression result}
The generalized Gaussian process demonstrates superior regression performance compared to the Gaussian process, particularly under the sparse observation condition, as shown in Fig.~\ref{fig:stepfunc-mean-covariance}. In the sparse observation scenario (left plot), the Gaussian process model produces smooth predictions that fail to capture the abrupt changes in the true function. This limitation arises from the reliance on the smooth nature of the squared exponential kernel function~\cite{williams2006gaussian}, which is not well-suited to modeling the sharp transitions. In contrast, the generalized Gaussian process model accurately captures these discontinuities by learning the mean and kernel functions directly from the data, allowing it to handle intricate data features more effectively. With a more extensive dataset (right plot), both models exhibit improved accuracy. From a Bayesian perspective, as the data volume increases, the posterior distribution becomes more influenced by the data and less by the prior assumptions. For the Gaussian process model, this translates into a better ability to capture the true underlying function, including complex features such as abrupt changes. However, despite the Gaussian process model's enhanced performance with dense observations, the generalized Gaussian process model retains a certain advantage, owing to its training on the same dataset, which enhances its capacity to adapt to intricate data features.

\subsubsection{Regression on a two-dimensional multi-scale trigonometric function}
\label{sec:results-gGP-trifunc}

\paragraph{Target function} 
Multi-scale functions present another challenging case due to their inherent complexity, characterized by varying length scales and non-stationarity with distinct properties across different regions. Traditional Gaussian processes often struggle to accurately capture these features because the employed kernel functions typically assume a fixed length scale and stationarity over the entire input space. As a specific example, the two-dimensional multi-scale trigonometric function used in this study is defined as follows:
\begin{equation}
\begin{aligned}
    f(x_1, x_2) &= \cos(2\pi x_1)\cos(2\pi x_2) + \sin(4\pi x_1)\sin(4\pi x_2) \mathds{1}_{[1/4,3/4]^2}(x_1, x_2) \\
    &+ \sin(8\pi x_1)\sin(8\pi x_2) \mathds{1}_{[1/2,3/4]^2}(x_1, x_2) + \sin(16\pi x_1)\sin(16\pi x_2) \mathds{1}_{[1/4,1/2]^2}(x_1, x_2).
\end{aligned}
\label{eq:2D-trifunc}
\end{equation}
The input space for $\bm{x} = (x_1, x_2)$ is defined as $[0,1]^2$. The two-dimensional indicator functions, such as $\mathds{1}_{[1/4,3/4]^2}(x_1, x_2)$, activate specific components of the function only within defined regions of the input space. For instance, the term $\sin(4\pi x_1)\sin(4\pi x_2)$ is only present when $(x_1, x_2)$ falls within the square region where both coordinates are between 1/4 and 3/4. This function incorporates four distinct length scales---1, 1/2, 1/4, and 1/8---and displays non-stationarity through its varying behaviors across different regions. These complexities require regression models to adapt to both changes in scale and local data structure, making this function an ideal benchmark for assessing a model's ability to handle intricate and heterogeneous data patterns.

\paragraph{Experiment setup} The objective is to predict the function values at 16,384 uniformly distributed test points ($2^{14}$ total, with $2^7$ along each dimension) within the input space using the provided observation data. The observation data $\mathcal{D}$ comprises input values $X = \left[\bm{x}^\text{dat}_1, \ldots, \bm{x}^\text{dat}_M\right]$ uniformly sampled from the input space, with corresponding target values $Y = \left[\tilde{y}^\text{dat}_1, \ldots, \tilde{y}^\text{dat}_M\right]^\top$ assumed to be corrupted by Gaussian noise:
\begin{equation}
    \tilde{y}_i^\text{dat} = f(\bm{x}_i^\text{dat}) + \epsilon_i,
\end{equation}
where $\epsilon_i \sim \mathcal{N}(0, 0.01^2)$. We consider two observation conditions: a sparse scenario with 256 observation points ($M = 256$, with 16 along each dimension) and a dense scenario with 1,024 observation points ($M=1024$, with 32 along each dimension). The number of observation points in the sparse scenario is determined based on the Nyquist--Shannon sampling theorem, which states that the sampling rate must be at least twice the highest frequency component present in the function to accurately reconstruct the signal. For the given multi-scale function, the highest frequency component corresponds to the finest scale of 1/8, necessitating a minimum sampling rate of 16 samples per unit length to avoid aliasing. Consequently, with 16 samples per dimension, the sparse observation grid is designed to meet this criterion, ensuring that the observed data sufficiently captures the essential features of the underlying function.

For each observation condition, both Gaussian process and generalized Gaussian process are employed to perform the regression task. In the Gaussian process approach, a squared exponential kernel, $\mathsf{K}(\bm{x}, \bm{x}') = \sigma_f^2 \exp\left[-\frac{1}{2} \left((x_1 - x_1')^2/\ell_1^2 + (x_2 - x_2')^2/\ell_2^2 \right) \right]$, is utilized. The hyperparameters, including the length scales $\ell_1$ and $\ell_2$, and the variance $\sigma_f$, are optimized by maximizing the log marginal likelihood. In the generalized Gaussian process approach, the neural operator is trained on the observation data, which involves optimizing the same likelihood-based criterion to ensure accurate function approximation.

\paragraph{Data generation} 
To train the generalized Gaussian process, we define the task as predicting a randomly selected subset $\cD' = \{X', Y'\}$ from the entire dataset $\cD$, conditioned on the remaining set $\cD \backslash \cD'$. Specifically, the training data consists of input-output pairs, where the input includes the test points $X'$ and the observations from $\cD \backslash \cD'$, and the output corresponds to the observed target values $Y'$. The set $X'$ contains $M'(< M)$ test points. In both the sparse and dense observation scenarios, $M'$ is set to 32, with the remaining 224 and 992 points used as observed points, respectively. As such, we randomly sample $5 \times 10^4$ data pairs without replacement for training in each scenario. With the training dataset, we train the generalized Gaussian process by minimizing the negative conditional log probability, as shown in Eq.~\eqref{eq:gGP-optimize}.

\begin{figure}[!htb]
    \centering
    \includegraphics[width=0.9\linewidth]{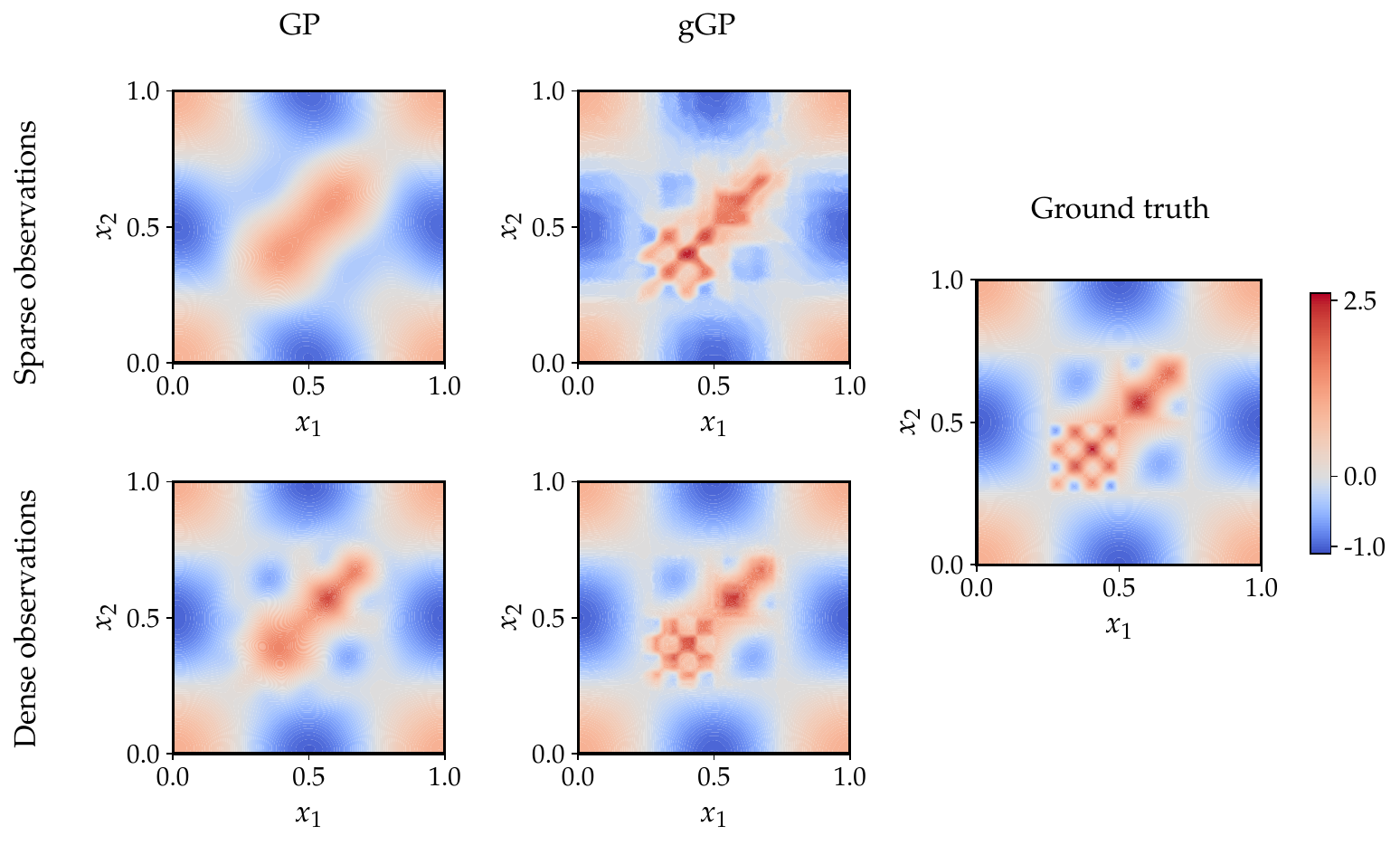}
    \caption{Comparison of regression performance for the two-dimensional multi-scale trigonometric function using Gaussian process (GP) and generalized Gaussian process (gGP) models. The top row displays the results with sparse observations on a $16 \times 16$ uniform grid, while the bottom row presents the results with dense observations on a $32 \times 32$ uniform grid. The GP captures, at best, the first two or three length scales in both scenarios, whereas the gGP consistently captures all four scales with higher precision.}
    \label{fig:trifunc-mean}
\end{figure}

\paragraph{Regression result}
The generalized Gaussian process model also demonstrates superior regression performance on the multi-scale function compared to the Gaussian process model, as shown in Fig.~\ref{fig:trifunc-mean}. With sparse observations (top row), the Gaussian process model (left plot) produces overly smooth predictions that fail to capture the intricate multi-scale features present in the ground truth (right plot). This limitation arises from its reliance on fixed length scales in the kernel function, which restricts its ability to adapt to varying scales within the data. Consequently, the Gaussian process model struggles to capture finer-scale details and variations inherent in the multi-scale function accurately. In contrast, the generalized Gaussian process model (middle plot) shows markedly better performance in capturing the multi-scale nature of the data, even with sparse observations. This improvement is attributed to the model's flexibility in learning length scales according to the data characteristics, enabling it to more accurately represent different scales and capture subtle variations that the Gaussian process model misses. As observation density increases (bottom row), both models show improved accuracy. However, the Gaussian process model remains unable to capture fine-scale details accurately, even with dense observations spaced at a quarter of the smallest scale. In contrast, the generalized Gaussian process model nearly captures all scales and demonstrates better predictive performance.

\begin{figure}[!htb]
    \centering
    \includegraphics[width=0.99\linewidth]{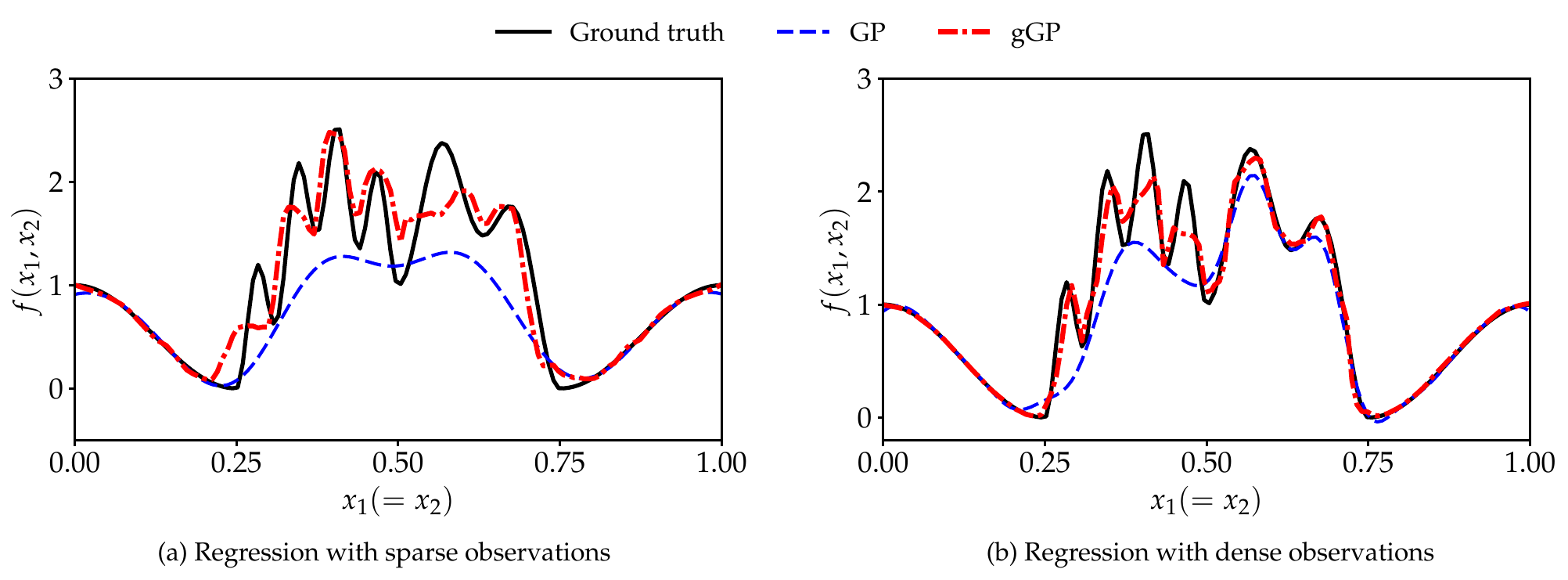}
    \caption{Comparison of regression results for the two-dimensional multi-scale trigonometric function along the diagonal ($x_1=x_2$) of the input domain using the Gaussian process (GP) and generalized Gaussian process (gGP): (a) predictions with sparse observations and (b) predictions with dense observations. The GP captures the first two scales with sparse observations and up to the third scale when using dense observations. In contrast, the gGP consistently captures all four scales, closely aligning with the ground truth.}
    \label{fig:trifunc-diag-mean-correlation}
\end{figure}

The comparison is more clearly illustrated in Fig.~\ref{fig:trifunc-diag-mean-correlation}, which presents the predicted function values along the diagonal of the input domain, specifically where $x_1 = x_2$. In the case of sparse observations (left plot), the Gaussian process model can only capture the first and second scales, but misses finer details. In contrast, the generalized Gaussian process model performs significantly better, closely aligning with the ground truth by accurately representing nearly all present scales. This trend continues with dense observations (right plot), where the Gaussian process model improves slightly by capturing up to the third scale but still lacks precision in representing finest-scale variations. The generalized Gaussian process model, however, consistently captures all four scales with high accuracy, demonstrating its superior ability to model complex multi-scale features in the data. This comparison clearly highlights the enhanced flexibility and accuracy of the generalized Gaussian process model compared to the Gaussian process model in handling multi-scale functions.

\subsection{Ensemble neural filter (EnNF) for sequential data assimilation}
\label{sec:results-EnNF}
We numerically investigate the assimilation performance of the ensemble neural filter using two chaotic dynamical models: the Lorenz-63 and the Lorenz-96 models, which are commonly used to validate data assimilation algorithms~\cite{spantini2022coupling}. Our results demonstrate that the ensemble neural filter can mimic the filtering performance of the ensemble Kalman filter when trained on data generated by the latter. Moreover, it exhibits excellent assimilation performance, even with an ensemble size as small as two, and maintains consistent performance across various small ensemble sizes, outperforming the ensemble Kalman filter method. The experimental setups for the Lorenz-63 and Lorenz-96 models primarily follow those described in the literature~\cite{strofer2021dafi} and~\cite{ahmed2020pyda}, respectively.

\subsubsection{Assimilation experiment on the Lorenz-63 model}
\label{sec:results-EnNF-63}

\paragraph{Dynamical model} The Lorenz-63 model, introduced by Lorenz~\cite{lorenz1963deterministic}, is a simplified mathematical model for describing atmospheric convection dynamics. The state vector in this model consists of three variables, represented as $\mathbf{z}(t) = \left(z_1(t), z_2(t), z_3(t)\right)$, whose dynamics are governed by the following set of three coupled ordinary differential equations (ODEs):
\begin{equation}
\begin{aligned}
& \frac{\mathrm{d} z_1}{\mathrm{~d} t}=\sigma(z_2 - z_1), \\
& \frac{\mathrm{d} z_2}{\mathrm{~d} t}=z_1 (\rho - z_3) - z_2, \\
& \frac{\mathrm{d} z_3}{\mathrm{~d} t}=z_1 z_2-\beta z_3,
\end{aligned}
\label{eq:Lorenz-63}
\end{equation}
where $\beta = 8/3$, $\rho = 28$, and $\sigma = 10$ are fixed model parameters commonly used to produce chaotic solutions towards the well-known Lorenz attractor. This ODE system is integrated using a fourth-order explicit Runge--Kutta method, with a fixed step size of $\Delta t = 0.01$. For twin experiment testing, we set the baseline initial condition slightly deviated from the true initial condition to evaluate the performance of data assimilation methods in correcting state estimates under initial condition uncertainties. 

\paragraph{Likelihood model}
The state of the Lorenz-63 system is observed directly every $\Delta t_\text{obs} = 0.5$ time units (i.e., every 50 integration steps). For any observation step $k$, the likelihood model is defined by:
\begin{equation}
\mathbf{d}(t_k) = \mathsf{H} \mathbf{z}^* (t_k) + \bm{\epsilon}(t_k),
\label{eq:likelihood}
\end{equation}
where $t_k$ is the $k$-th observation time given by $t_k = k \Delta t_\text{obs}$, $\mathbf{z}^*$ represents the true solution, $\mathbf{d}$ denotes the observation, and $\bm{\epsilon}$ represents the observational noise. In this experiment, $\mathsf{H} \in \mathbb{R}^{2 \times 3}$ is a linear observation operator that selects the state variables $z_1$ and $z_3$ from the full state vector, and $\bm{\epsilon}$ is assumed to be Gaussian with zero mean and covariance $\mathbf{C}_{d d, i, i} = \left(0.1\,\mathbf{d}^*_i + 0.05\right)^2$ for $i \in \{1, 2\}$, where the synthetic observation $\mathbf{d}^* = \mathsf{H} \mathbf{z}^*$ and $\mathbf{d}^*_i$ refers to its $i$-th component.

\paragraph{Experiment setup}
Given the true initial condition of $z_1(0) = -8.5, z_2(0) = -7.0, z_3(0) = 27.0$, we generate the true solution and a sequence of synthetic observations. The baseline initial condition is set slightly off from the true initial condition at $z_1(0) = -8.0, z_2(0) = -9.0, z_3(0) = 28.0$. The initial prior ensemble, $\mathbf{Z}^{\mathrm{f}} (0)$, is generated by sampling from a Gaussian distribution with mean equal to the baseline initial condition and covariance $\mathbf{C}_{z z} = \operatorname{diag}(0.4,2.0,1.4)$.

With the initial prior ensemble generated, the dynamical model available, and both the synthetic observations and likelihood model defined, the ensemble Kalman filter is applied to correct the state estimates whenever an observation is made. Specifically, during each assimilation window, each realization $\mathbf{z}_i$ is independently propagated according to Eq.~\eqref{eq:Lorenz-63}, and then updated based on the ensemble statistics and its corresponding perturbed observation $\mathbf{d}_i$, as described by Eq.~\eqref{eq:EnKF-update}. The updated ensemble then serves as the starting point for the next data assimilation window. We implement the ensemble Kalman filter using an ensemble size of $N = 50$ to ensure that the calculated ensemble error covariances closely approximate the true error covariances. 

\paragraph{Data generation}
We conduct ten separate runs of the ensemble Kalman filter on the Lorenz-63 model to generate data for training the ensemble neural filter. Each run begins with a different initial prior ensemble and spans a training period of $T = 75$ time units that covers 150 assimilation windows. As such, this process generates 1,500 pairs of prior and posterior ensembles of full state vectors for training purposes. However, as stated above, we assume each state variable in a prior ensemble shares the same update operator, which allows its independent update without involving other state variables. Thus, we have 4,500 input-output data pairs for training ensemble neural filter, given the state vector has three variables. Specifically, the input is an ensemble consisting of three components: a prior state variable, predicted observation, and observation, i.e., $\left\{\left(z_{i,j}^\mathrm{f},\mathsf{H}\mathbf{z}_i^\mathrm{f}, \mathbf{d}_i\right)\right\}_{i=1}^{50}$ for $j \in \{1, 2, 3\}$; the output is an ensemble of the corresponding posterior state variable, i.e., $\left\{z_{i,j}^\mathrm{a}\right\}_{i=1}^{50}$.
With the training dataset, we train the ensemble neural filter on an NVIDIA V100 GPU for 2,000 epochs using the open-source machine learning framework PyTorch~\cite{paszke2019pytorch}.

\begin{figure}[!htb]
    \centering
    \includegraphics[width=0.99\textwidth]{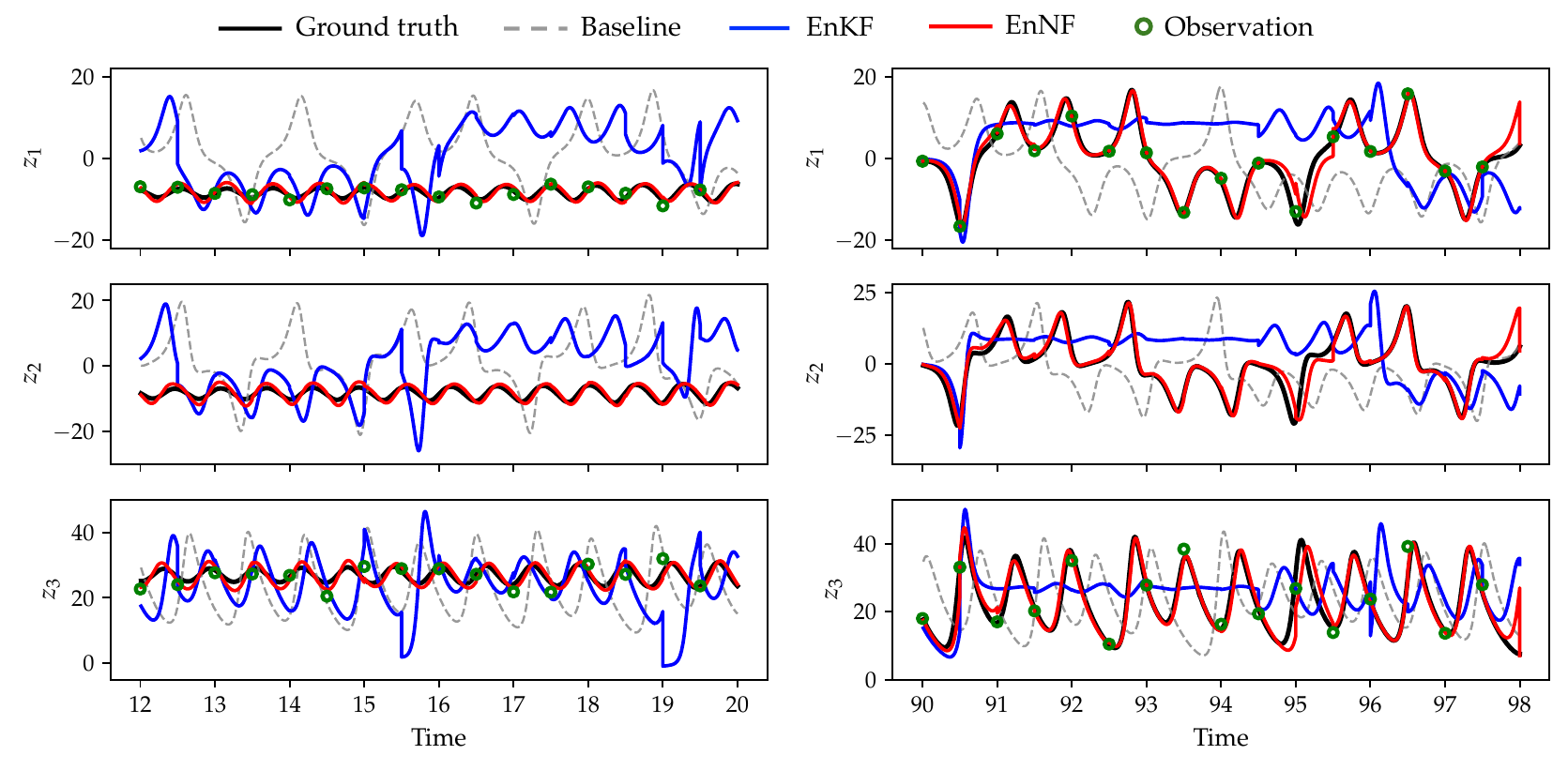}
    \caption{Comparison of state estimation for the Lorenz-63 model using ensemble Kalman filter (EnKF) and ensemble neural filter (EnNF) with the smallest ensemble size of $N = 2$. The left panels show the dynamics of three state variables, $z_1$, $z_2$, and $z_3$, over 16 initial assimilation windows, while the right panels continue the display over another 16 windows after the process has been going on for a while. Each panel compares the results of the baseline, EnKF and EnNF against the ground truth. The baseline result deviates significantly from the ground truth due to the initial condition deviation and the absence of data assimilation. The EnKF result shows limited improvement due to the small ensemble size. In contrast, the EnNF result is very close to the ground truth, demonstrating its superior assimilation performance. Note that the results for both EnKF and EnNF represent the behavior of the ensemble mean $\overline{\mathbf{z}} (t)$.}
    \label{fig:Lorenz-63-compare}
\end{figure}

\paragraph{Assimilation result}
The trained ensemble neural filter exhibits excellent assimilation performance, even with the smallest ensemble size of $N = 2$, as shown in Fig.~\ref{fig:Lorenz-63-compare}. This figure illustrates the dynamics of each state variable in the ensemble mean, $\overline{\mathbf{z}} (t) :=\frac{1}{N} \sum_{i=1}^N \mathbf{z}_i (t)$, across two different time periods, each spanning 16 assimilation windows. The first period (left column) corresponds to the initial phase within the training period, while the second period (right column) falls within the forecasting period after the training period and represents temporal extrapolation at a later stage. It can be seen that, in both periods, the baseline dynamics of each state variable significantly deviates from the ground truth due to the minor initial condition deviations and the chaotic nature of the Lorenz-63 system. Despite the observations, the ensemble Kalman filter fails to correct the baseline state effectively; sometimes updating the state even further from the observations or resulting in minimal updates. This is attributed to the small ensemble size, which results in erroneous approximations of the true covariances, an important component in the ensemble Kalman filter update equation. In contrast, the results from ensemble neural filter closely match the ground truths, as it bypasses the need for explicit covariance approximations.

\begin{figure}[!htb]
    \centering
    \includegraphics[width=0.99\textwidth]{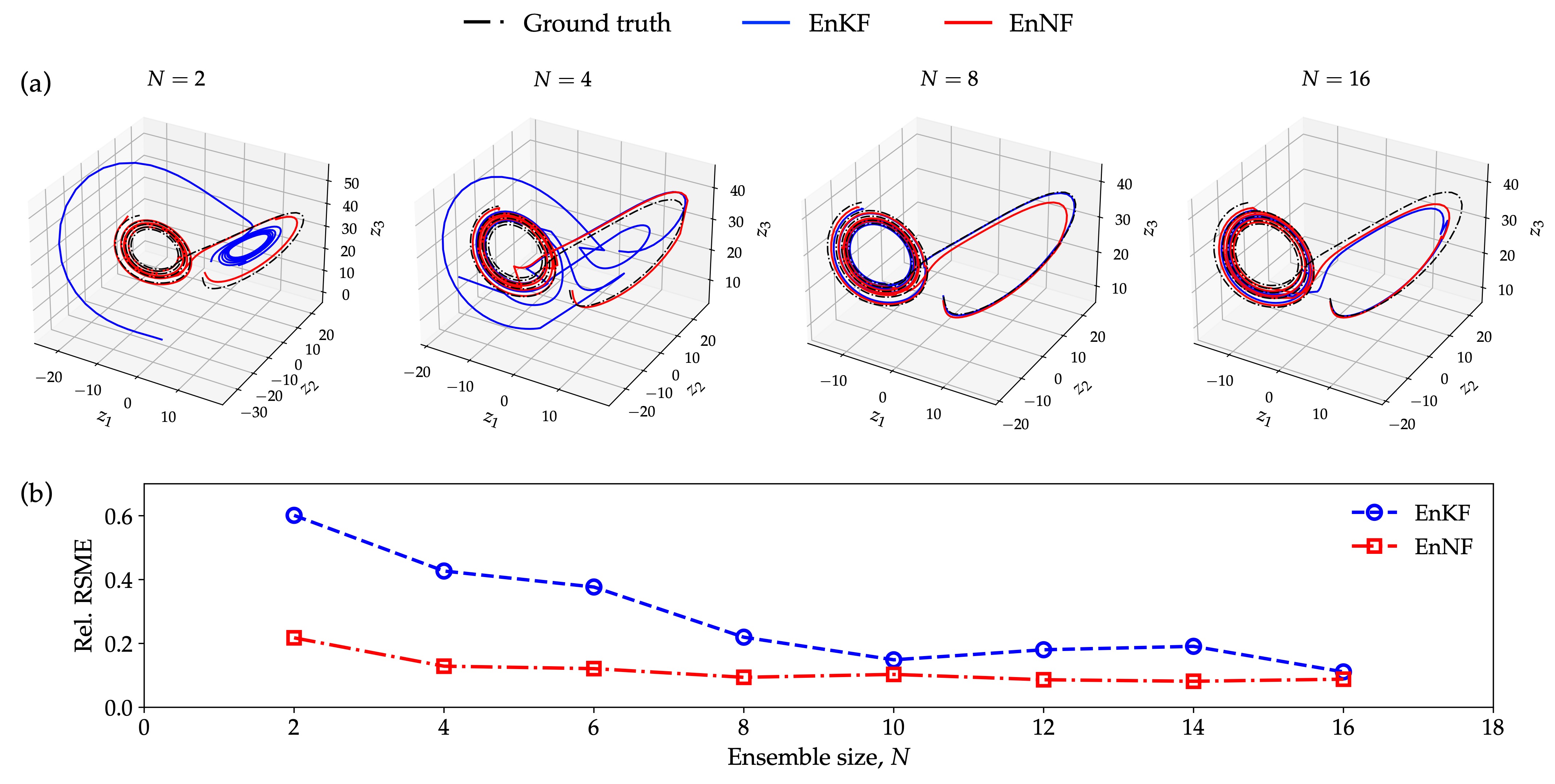}
    \caption{Comparison of assimilation performance of ensemble Kalman filter (EnKF) and ensemble neural filter (EnNF) for the Lorenz-63 model across various ensemble sizes. Panel (a) shows the ensemble mean trajectories using EnKF and EnNF under four ensemble sizes over 20 assimilation windows during the process. The trajectory from EnKF gradually converges towards the ground truth with increasing ensemble size, while the trajectory from EnNF consistently remains in close proximity to the ground truth. Panel (b) quantitatively compares the assimilation errors of both methods across eight ensemble sizes from 2 to 16. The EnNF consistently demonstrates lower errors compared to EnKF, highlighting its superior assimilation performance and robustness, especially with smaller ensemble sizes.}
    \label{fig:Lorenz-63-diff-size}
\end{figure}

Moreover, the trained ensemble neural filter demonstrates consistent assimilation performance across various small ensemble sizes, as illustrated in Fig.~\ref{fig:Lorenz-63-diff-size}. Specifically, Fig.~\ref{fig:Lorenz-63-diff-size}(a) compares the model trajectories using the ensemble Kalman filter and ensemble neural filter against the ground truths across four different ensemble sizes: $N = 2, 4, 8$ and 16. It can be seen that, as the ensemble size increases, the trajectory from ensemble Kalman filter noticeably converges towards the ground truth, indicating improved assimilation performance with larger ensembles. In contrast, the trajectory of ensemble neural filter consistently aligns closely with the ground truth across all four ensemble sizes, highlighting its robustness under varying uncertainty conditions. This superior assimilation performance is further evidenced in Fig.~\ref{fig:Lorenz-63-diff-size}(b), which shows the assimilation errors, represented by relative root mean square error (RMSE), for both methods across eight ensemble sizes ranging from 2 to 16. The relative RMSE is computed following the equation:
\begin{equation}
\text{relative}\ \mathrm{RMSE}=\sqrt{\frac{\sum_{k=1}^{N_a}\left\|\overline{\mathbf{z}}^{\mathrm{a}} (t_k)- \mathbf{z}^* (t_k)\right\|^2}{\sum_{k=1}^{N_a}\left\|\mathbf{z}^* (t_k)\right\|^2}},
\label{eq:RMSE}
\end{equation}
where $N_a$ represents the total number of assimilation windows, $\|\cdot\|$ indicates the $\ell^2$-norm, and $\overline{\mathbf{z}}^{\mathrm{a}}$ is the posterior ensemble mean obtained by either ensemble Kalman filter or ensemble neural filter. Here, the assimilation error is calculated over a span of 200 assimilation windows, including the initial 150 in the training period and the subsequent 50 in the extrapolation period. For each setting, defined by a specific assimilation method and ensemble size, the error represents the average metric from five runs with different initial priors, ensuring statistical robustness and reliability. It is clear that the assimilation errors from ensemble Kalman filter are initially high with very small ensembles but decrease rapidly as the ensemble size increases. In contrast, ensemble neural filter maintains lower values across all ensemble sizes, further demonstrating its superior assimilation performance. Notably, the errors from ensemble neural filter remain stable and low throughout the entire interval, highlighting its effectiveness and robustness in handling smaller ensembles compared to the ensemble Kalman filter. This characteristic makes ensemble neural filter particularly valuable in scenarios where computational resources are limited.

\subsubsection{Assimilation experiment on the Lorenz-96 model}
\label{sec:results-EnNF-96}
\paragraph{Dynamical model}
The Lorenz-96 model is a widely used dynamical model developed to study fundamental issues related to mid-latitude atmospheric and weather predictability~\cite{lorenz1996predictability}. Unlike the Lorenz-63 model, which describes the fluid convection between hot and cold plates using three variables, the Lorenz-96 model extends to higher dimensions, making it a popular testbed for data assimilation algorithms in a more complex yet controlled environment. The state vector in this model consists of $m$ ($m \geq 4$) variables, represented as $\mathbf{z}(t)=\left(z_1(t), \ldots, z_m(t)\right)$, whose dynamics are given by a set of nonlinear ODEs:
\begin{equation}
\frac{\mathrm{d} z_i}{\mathrm{~d} t}=\left(z_{i+1}-z_{i-2}\right) z_{i-1}-z_i+F, \quad i=1,2, \ldots, m,
\label{eq:Lorenz-96}
\end{equation}
with periodic boundary conditions $z_{-1} = z_{m-1}$, $z_0 = z_m$, and $z_{m+1} = z_1$. Here, $m$ is the number of spatial points in the discretization, $z_i$ denotes the state variable at the $i$-th location, and $F$ represents a constant forcing parameter. In this experiment, we use $n = 24$ and $F = 8$, a commonly used configuration to induce fully chaotic dynamics. The well-defined ODE system is integrated using a fourth-order explicit Runge--Kutta method, with a fixed step size of $\Delta t = 0.01$.

\paragraph{Likelihood model}
The state of the Lorenz-96 system is observed directly every $\Delta t_\text{obs} = 0.2$ time units (i.e., every 20 integration steps), considering its increased complexity compared to the Lorenz-63 model. The likelihood model follows the same form as that used for the Lorenz-63 system. Here, we recall it for completeness:
\begin{equation*}
\mathbf{d}(t_k) = \mathsf{H} \mathbf{z}^* (t_k) + \bm{\epsilon}(t_k).
\end{equation*}
For the Lorenz-96 model, the linear observation operator $\mathsf{H} \in \mathbb{R}^{8 \times 24}$ selects the state variables at eight equidistant locations from the full state vector, specifically $z_i$ with $i = 3, 6, \ldots, 24$. The Gaussian observational noise $\bm{\epsilon}$ has zero mean and covariance $\mathbf{C}_{d d, i, i} = \left(0.05\,\mathbf{d}^*_i + 0.1\right)^2$ for $i \in \{1, 2, \ldots, 8\}$.

\paragraph{Experiment setup}
To generate a valid initial condition, we start at $t = -5$ with the equilibrium state $z_i = F$ for $i = 1, \ldots, 24$, and introduce a small perturbation to the $20$th state variable by setting $z_{20} = F + 0.01$. We then run the ODE integrator up to $t = 0$ and use the solution at $t = 0$ as the true initial condition. With this true initial condition, we generate the true solution and a sequence of synthetic observations. For twin experiment testing, the baseline initial condition is set slightly off from the true initial condition by adding a random noise drawn from $\mathcal{N}\left(\mathbf{0}, \mathbf{I}_{24}\right)$. The initial prior ensemble, $\mathbf{Z}^{\mathrm{f}} (0)$, is then generated by sampling from a Gaussian distribution with mean equal to the baseline initial condition and covariance $\mathbf{C}_{z z}=\mathbf{I}_{24}$.

With the initial prior ensemble generated, the dynamical model available, and both the synthetic observations and likelihood model specified, we apply the ensemble Kalman filter for state estimate under initial condition uncertainties, following a similar procedure as described for the Lorenz-63 model. For the Lorenz-96 model, we use an ensemble size of $N = 100$, given its significantly higher dimensionality. Note that this ensemble size does not result in optimal assimilation performance according to the literature~\cite{spantini2022coupling}; however, we use this size for proof of concept and computational manageability.

\paragraph{Data generation}
We conduct 30 separate runs of the ensemble Kalman filter on the Lorenz-96 model to generate data for training ensemble neural filter, with each run starting with a different initial prior ensemble and spanning a training period of $T = 30$ time units that covers 150 assimilation windows. This process generates 4,500 pairs of prior and posterior ensembles of full state vectors for training purposes. Given that the model state consists of 24 variables, and assuming each prior state variable shares the same update operator, we have 108,000 training data pairs, with the input and output represented as $\left\{\left(z_{i,j}^\mathrm{f},\mathsf{H}\mathbf{z}_i^\mathrm{f}, \mathbf{d}_i\right)\right\}_{i=1}^{100}$ and $\left\{z_{i,j}^\mathrm{a}\right\}_{i=1}^{100}$ for $j \in \{1, 2, \ldots, 24\}$, respectively.

However, unlike the training strategy used for the Lorenz-63 case, in this case, the ensemble neural filter is initially trained using the entire dataset for 2000 epochs and then fine-tuned for each state variable using their respective sub-datasets for a few hundred epochs. This adjustment is necessary because an ensemble neural filter with the same architecture, in terms of the number of layers and neurons, as used in the Lorenz-63 case may not sufficiently handle the filtering complexities here. Alternatively, training a larger-scale ensemble neural filter with more layers and neurons may directly address this issue.

\begin{figure}[!htb]
    \centering
    \includegraphics[width=0.99\textwidth]{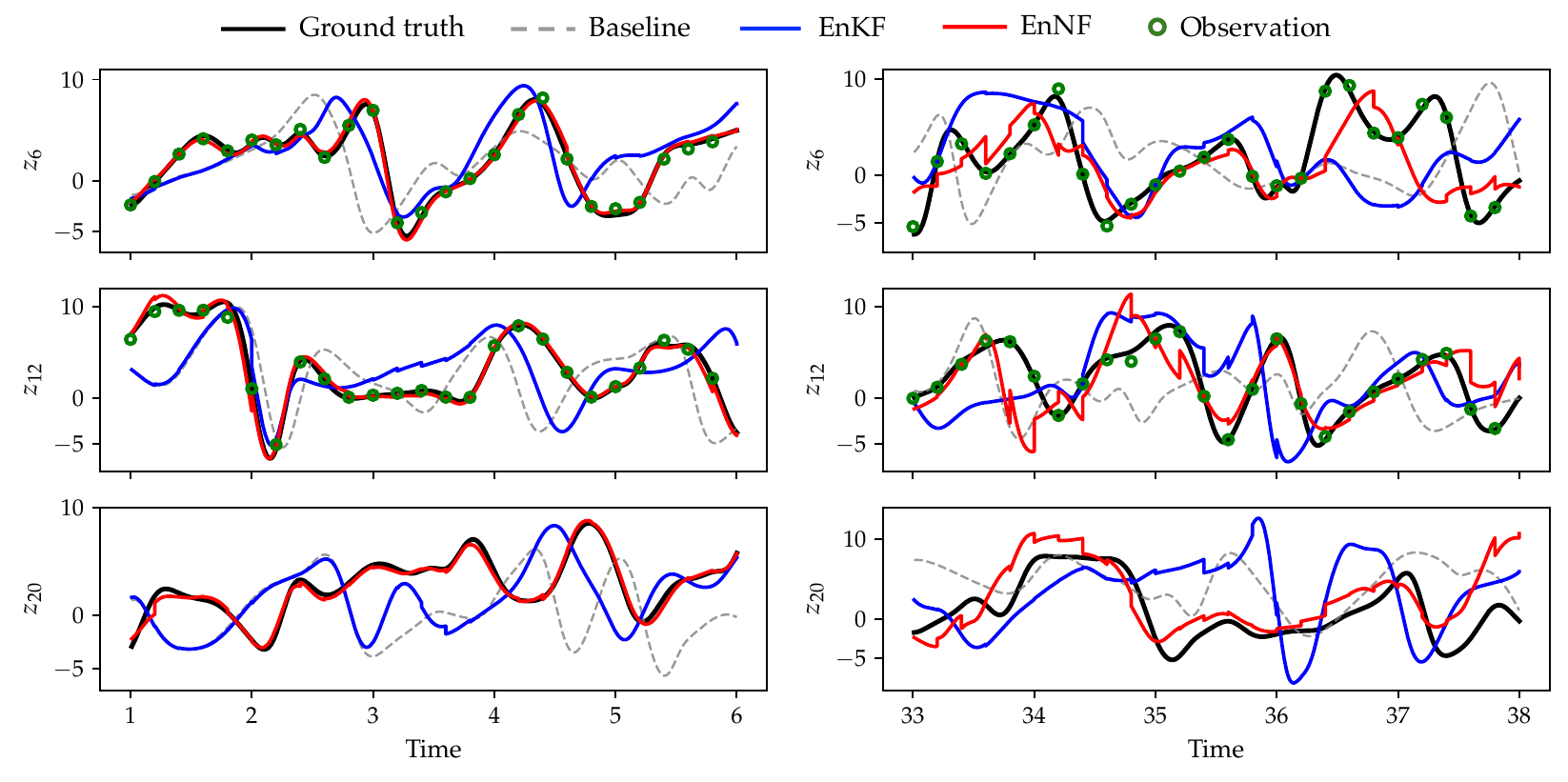}
    \caption{Comparison of state estimation for the Lorenz-96 model using ensemble Kalman filter (EnKF) and ensemble neural filter (EnNF) with the smallest ensemble size of $N = 2$. The left panels show the dynamics of three state variables, $z_6$, $z_{12}$, and $z_{20}$, over 25 initial assimilation windows, while the right panels continue the display over another 25 windows after the process has been going on for a while. Each panel compares the results of the baseline, EnKF and EnNF against the ground truth. The baseline result deviates evidently from the ground truth due to the initial condition deviation and the absence of data assimilation. The EnKF result shows limited improvement due to the small ensemble size. In contrast, the EnNF result is significantly closer to the ground truth, demonstrating its superior assimilation performance under this extremely small ensemble size. Note that the results for both EnKF and EnNF represent the dynamics of the three state variables from the ensemble mean $\overline{\mathbf{z}} (t)$.}
    \label{fig:Lorenz-96-compare}
\end{figure}

\paragraph{Assimilation result}
The trained ensemble neural filter also demonstrates good assimilation performance on this more complex model, even with the smallest ensemble size $N = 2$, as shown in Fig.~\ref{fig:Lorenz-96-compare}. Specifically, we compare the dynamics of three state variables, $z_6$, $z_{12}$, and $z_{20}$, from ensemble Kalman filter and ensemble neural filter against the ground truth over two distinct time periods, with the first period (left column) representing an initial phase within the training period and the other (right column) temporal extrapolation at a later stage, both covering 25 assimilation windows. Ensemble neural filter performs reasonably well under this extreme small-ensemble condition, while ensemble Kalman filter falls short. The ensemble mean from ensemble neural filter virtually matches the ground truth during the initial period and remains near the true solution in the forecasting period despite some discrepancies due to the limited training dataset. By employing a larger dataset collected over a sufficiently long period, the trained ensemble neural filter could potentially enhance its assimilation performance for temporal extrapolation.

\begin{figure}[!htb]
    \centering
    \includegraphics[width=0.9\textwidth]{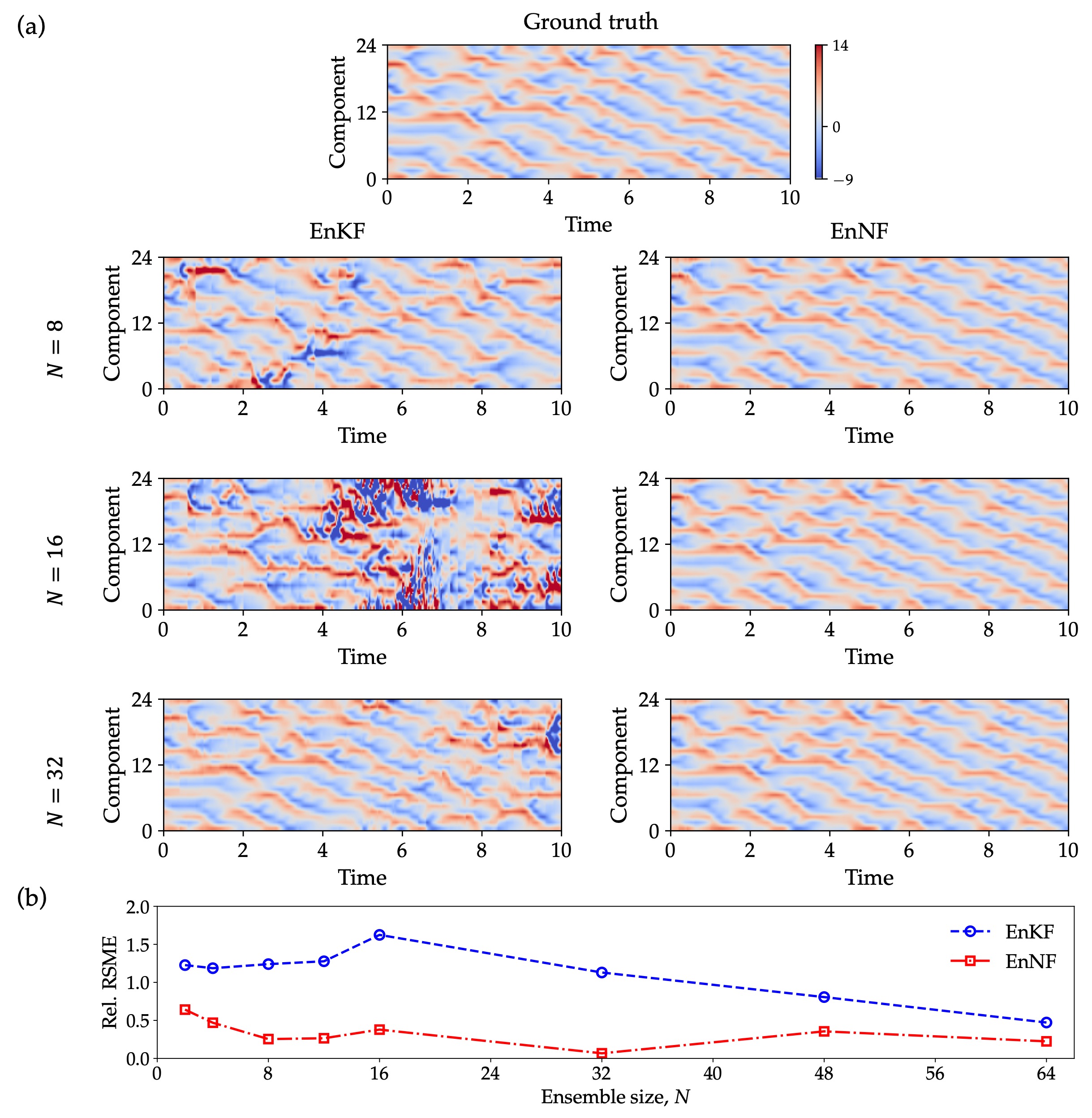}
    \caption{Comparison of assimilation performance of ensemble Kalman filter (EnKF) and ensemble neural filter (EnNF) for the Lorenz-96 model across various ensemble sizes. Panel (a) shows the traveling waves, i.e., ensemble mean of the system over time, from EnKF and EnNF under three ensemble sizes over the initial 50 assimilation windows. The waves from EnKF exhibit clear deviations from the ground truth, while those from EnNF show visual identity. Panel (b) quantitatively compares the assimilation errors of both methods across eight ensemble sizes ranging from 2 to 64. The EnNF consistently demonstrates lower errors, highlighting its superior assimilation performance and robustness for small ensemble sizes in the Lorenz-96 model.}
    \label{fig:Lorenz-96-diff-size}
\end{figure}

Additionally, the trained ensemble neural filter exhibits robust and consistent assimilation performance across various small ensemble sizes, which is illustrated in Fig.~\ref{fig:Lorenz-96-diff-size}. We compare the results of ensemble Kalman filter and ensemble neural filter against the ground truth using both qualitative and quantitative methods in Fig.\ref{fig:Lorenz-96-diff-size}(a) and (b), respectively. Fig.\ref{fig:Lorenz-96-diff-size}(a) shows the traveling waves in the Lorenz-96 model within the initial 50 assimilation windows by plotting the ensemble mean values (with 24 components) over time. The traveling waves from ensemble Kalman filter differ significantly from the ground truth for three ensemble sizes ($N = 8, 16$, and 32), with the $N = 16$ case showing clear extreme estimates. In contrast, the traveling waves from ensemble neural filter closely match the ground truth for all three ensemble sizes, demonstrating superior assimilation performance. Fig.\ref{fig:Lorenz-96-diff-size}(b) further illustrates this by comparing the assimilation errors for both methods across eight ensemble sizes ranging from 2 to 64. As with the Lorenz-63 model, the error is calculated over 200 assimilation windows---150 during the training period followed by 50 during the extrapolation period. Each assimilation error is averaged over five runs with different initial priors to ensure statistical robustness and reliability. For smaller ensemble sizes ($N \leq 16$), the assimilation errors for ensemble Kalman filter consistently exceed 100\%. Beyond this point, further increasing the ensemble size gradually reduces assimilation errors to a lower level. In contrast, the assimilation errors for ensemble neural filter remain stable and low across the entire range, highlighting its superior effectiveness and robustness in handling small ensembles compared to ensemble Kalman filter. 

\section{Conclusion}
\label{sec:conclude}
Approximate Bayesian inference is typically approached through deterministic and stochastic methods, each with significant limitations: deterministic methods often rely on predefined functional structures, while stochastic methods require large sample sizes for reliability. To address these challenges, we introduce the BI-EqNO, an equivariant neural operator framework for enhancing various approximate Bayesian inference methods. BI-EqNO transforms prior distributions into corresponding posterior distributions through data-driven training and ensures necessary permutation equivariance between their representations.

We have rigorously evaluated this framework in two important scenarios: regression and sequential data assimilation, demonstrating its efficacy and broad applicability. In regression, the generalized Gaussian process (gGP) outperforms traditional Gaussian process in learning functions characterized by abrupt changes and multiple scales. This enhanced performance is attributable to its ability to learn covariances directly from the data, free from the restriction of predefined kernel structures. In sequential data assimilation, the ensemble neural filter (EnNF) is trained to reproduce the filtering performance of the ensemble Kalman filter and demonstrates superior performance under small-ensemble settings in both Lorenz-63 and Lorenz-96 models.

Despite these promising results, further investigation is required to apply both the generalized Gaussian process and ensemble neural filter to complex, real-world scenarios. For the generalized Gaussian process, integrating essential physical constraints, such as frame independence, is crucial when modeling physical systems. Adapting such physics-informed generalized Gaussian process to real-world challenges, particularly in large-scale geophysical contexts with asymmetric and non-stationary behaviors, necessitates systematic exploration. Additionally, training the ensemble neural filter in high-dimensional systems presents substantial challenges due to significant memory demands. Advancements using modern machine learning architectures, such as Transformers~\cite{goel2023can, vaswani2017attention}, are essential for enhancing its scalability and performance.

\section*{Acknowledgments}

XHZ is supported by the U.S. Air Force under agreement number FA865019-2-2204. The U.S. Government is authorised to reproduce and distribute reprints for Governmental purposes notwithstanding any copyright notation thereon. ZRL and HX are funded by Deutsche Forschungsgemeinschaft (DFG, German Research Foundation) under Germany's Excellence Strategy - EXC 2075 – 390740016. We acknowledge the support by the Stuttgart Center for Simulation Science (SimTech). We appreciate the valuable discussions with members of SimTech, particularly Wolfgang Nowak, Steffen Staab, Tim Schneider, and Flex Frizen.

\appendix
\section{Mathematical constraints of generalized Gaussian process}
\label{app:gGP-constraints}
In designing the generalized Gaussian process, it is essential to preserve the mathematical constraints inherent in the conditional posterior distribution of Gaussian process regression. For completeness, we restate the mean and covariance:
\begin{equation}
\begin{aligned}
\mathsf{m}_\cD(X^*) & = \mathsf{m}(X^*) + \mathsf{K}(X^*, X)\left[\mathsf{K}(X, X)+\sigma_\epsilon^2 \mathbf{I}\right]^{-1} \left[Y - \mathsf{m}(X)\right], \\
\mathsf{K}_\cD(X^*, X^*) & =\mathsf{K}(X^*, X^*)-\mathsf{K}(X^*, X)\left[\mathsf{K}(X, X)+\sigma_\epsilon^2 \mathbf{I}\right]^{-1} \mathsf{K}(X, X^*).
\end{aligned}
\end{equation}
The equations suggest that the mean vector and covariance matrix are both permutation invariant to the observation data $(X, Y)$ and permutation equivariant to the test data points $X^*$. 

\paragraph{Permutation invariance to observation data}
To demonstrate that the mean vector $\mathsf{m}_\cD(X^*)$ is invariant under permutations of the observation data, consider a permutation $\mathcal{P}$ of the observation data. Let the reordered input and target values be $X' = \mathcal{P} X$ and $Y' = \mathcal{P} Y$. The covariance matrix and the cross-covariance with the test points will transform as:
\begin{equation}
    \mathsf{K}(X', X') = \mathcal{P} \mathsf{K}(X, X) \mathcal{P}^\top \quad \text{and} \quad
    \mathsf{K}(X^*, X') = \mathsf{K}(X^*, X) \mathcal{P}^\top.
\end{equation}
The posterior mean vector under this permutation is given by:
\begin{equation}
    \mathsf{m}'_\cD(X^*) = \mathsf{m}(X^*) + \mathsf{K}(X^*, X')\left[\mathsf{K}(X', X')+\sigma_\epsilon^2 \mathbf{I}\right]^{-1} \left[Y' - \mathsf{m}(X')\right].
\end{equation}
Substituting the transformed variables:
\begin{equation}
    \mathsf{m}'_\cD(X^*) = \mathsf{m}(X^*) + \mathsf{K}(X^*, X)\mathcal{P}^\top\left[\mathcal{P} \mathsf{K}(X, X) \mathcal{P}^\top + \sigma_\epsilon^2 \mathbf{I}\right]^{-1} \left[\mathcal{P}Y - \mathsf{m}(\mathcal{P}X)\right].
\end{equation}
Using the orthogonality of permutation matrices ($\mathcal{P}^{-1} = \mathcal{P}^\top$):
\begin{equation}
    \mathsf{m}'_\cD(X^*) = \mathsf{m}(X^*) + \mathsf{K}(X^*, X)\mathcal{P}^\top \mathcal{P}\left[ \mathsf{K}(X, X) + \sigma_\epsilon^2 \mathbf{I}\right]^{-1}  \mathcal{P}^\top \mathcal{P} \left[Y - \mathsf{m}(X)\right].
\end{equation}
Substituting $\mathcal{P}^\top \mathcal{P} = \mathbf{I}$ into the expression:
\begin{equation}
    \mathsf{m}'_\cD(X^*) = \mathsf{m}(X^*) + \mathsf{K}(X^*, X)\left[ \mathsf{K}(X, X) + \sigma_\epsilon^2 \mathbf{I}\right]^{-1} \left[Y - \mathsf{m}(X)\right].
\end{equation}
Thus, $\mathsf{m}'_\cD(X^*) = \mathsf{m}_\cD(X^*)$. This confirms that the posterior mean vector is invariant under permutations of the observation data.

\paragraph{Permutation equivariance to test points}
To demonstrate that the mean vector $\mathsf{m}_\cD(X^*)$ is equivariant under permutations of the test points, consider a permutation $\mathcal{Q}$ of the test points. Let the reordered test points be ${X^*}' = \mathcal{Q} X^*$. The cross-covariance with the observation points will transform as:
\begin{equation}
    \mathsf{K}({X^*}', X) = \mathcal{Q} \mathsf{K}(X^*, X).
\end{equation}
The posterior mean vector under this permutation is given by:
\begin{equation}
    \mathsf{m}_\cD({X^*}') = \mathsf{m}({X^*}') + \mathsf{K}({X^*}', X)\left[\mathsf{K}(X, X)+\sigma_\epsilon^2 \mathbf{I}\right]^{-1} \left[Y - \mathsf{m}(X)\right].
\end{equation}
Substituting the transformed variables:
\begin{equation}
    \mathsf{m}_\cD({X^*}') = \mathsf{m}(\mathcal{Q} X^*) + \mathcal{Q} \mathsf{K}(X^*, X)\left[\mathsf{K}(X, X)+\sigma_\epsilon^2 \mathbf{I}\right]^{-1} \left[Y - \mathsf{m}(X)\right].
\end{equation}
Since $\mathsf{m}(\mathcal{Q} X^*)$ is the reordered prior mean vector, and $\mathcal{Q} \mathsf{K}(X^*, X)$ represents the reordered cross-covariance matrix:
\begin{equation}
\begin{aligned}
    \mathsf{m}_\cD({X^*}') &= \mathcal{Q}\mathsf{m}(X^*) + \mathcal{Q} \mathsf{K}(X^*, X)\left[\mathsf{K}(X, X)+\sigma_\epsilon^2 \mathbf{I}\right]^{-1} \left[Y - \mathsf{m}(X)\right] \\ &= \mathcal{Q} \left[\mathsf{m}(X^*) + \mathsf{K}(X^*, X)\left[\mathsf{K}(X, X)+\sigma_\epsilon^2 \mathbf{I}\right]^{-1} \left[Y - \mathsf{m}(X)\right] \right]
\end{aligned}
\end{equation}
Thus, $\mathsf{m}_\cD({X^*}') = \mathcal{Q} \mathsf{m}_\cD({X^*})$. This demonstrates that the posterior mean vector is equivariant with respect to the permutation $\mathcal{Q}$ of the test points.

Similarly, it can be shown that the covariance matrix $\mathsf{K}_\cD(X^*, X^*)$ is permutation invariant to the observation data and permutation equivariant to the test data points. This can be demonstrated as follows:
\begin{enumerate}[(1)]
    \item Under reordered observations, $X' = \mathcal{P} X$ and $Y' = \mathcal{P} Y$, the covariance matrix remains invariant:
    \begin{equation}
    \mathsf{K}'_\cD(X^*, X^*) = \mathsf{K}_\cD(X^*, X^*).
    \end{equation}
    \item Under reordered test points, ${X^*}' = \mathcal{Q} X^*$, the covariance matrix transforms equivariantly:
    \begin{equation}
    \mathsf{K}_\cD({X^*}', {X^*}') = \mathcal{Q} \mathsf{K}_\cD(X^*, X^*) \mathcal{Q}^\top.
\end{equation}
\end{enumerate}

\section{Mathematical properties of ensemble neural filter}
\label{app:EnNF-properties}

In designing the ensemble neural filter (EnNF), we preserve two fundamental properties from the ensemble Kalman filter (EnKF) to ensure efficient training and application:
\begin{enumerate}
    \item Uniform update operation for each state variable.
    \item The update relies solely on information from the observation space.
\end{enumerate}

In the ensemble Kalman filter framework, updated ensemble realizations are confined to the subspace spanned by the prior ensemble~\cite{evensen2022data}. Specifically, an updated ensemble realization is a linear combination of prior ones, expressed in matrix form as:
\begin{equation}
    \mathbf{Z}^\text{a} =\mathbf{Z}^\text{f} + \mathbf{Z}^\text{f} \mathbf{W},
    \label{eq:EnKF_linear_update}
\end{equation}
where $\mathbf{Z}^{\mathrm{f}}=\left(\mathbf{z}_1^{\mathrm{f}}, \ldots, \mathbf{z}_N^{\mathrm{f}}\right) \in \mathbb{R}^{n \times N}$ denotes the prior ensemble, $\mathbf{Z}^{\mathrm{a}}=\left(\mathbf{z}_1^{\mathrm{a}}, \ldots, \mathbf{z}_N^{\mathrm{a}}\right) \in \mathbb{R}^{n \times N}$ represents the posterior ensemble, and $\mathbf{W}\in \mathbb{R}^{N \times N}$ is the weight matrix. 

Consider the $j$-th state variable across all realizations. We define the vector of the 
$j$-th state variable from the prior ensemble as:
\begin{equation}
\mathbf{z}_{:, j}^\mathrm{f}=\left(z_{1, j}^\mathrm{f}, z_{2, j}^\mathrm{f}, \ldots, z_{N, j}^\mathrm{f}\right),
\end{equation}
and similarly for the posterior ensemble:
\begin{equation}
\mathbf{z}_{:, j}^\mathrm{a}=\left(z_{1, j}^\mathrm{a}, z_{2, j}^\mathrm{a}, \ldots, z_{N, j}^\mathrm{a}\right).
\end{equation}
From the linear update scheme~\eqref{eq:EnKF_linear_update}, the update for the $j$-th state variable across all realizations is:
\begin{equation}
    \mathbf{z}_{:,j}^{\mathrm{a}} = \mathbf{z}_{:,j}^{\mathrm{f}} + \mathbf{z}_{:,j}^{\mathrm{f}} \mathbf{W} \quad \text{for} \quad j \in \{1, \ldots, n\},
\end{equation}
indicating that each state variable follows the same update mechanism.

The weight matrix $\mathbf{W}$ is determined by the following equation derived from Eq.~\eqref{eq:EnKF-update}:
\begin{equation}
    \mathbf{W} = \mathbf{Y}^\top \left(\overline{\mathbf{C}}_{y y} + \overline{\mathbf{C}}_{d d}\right)^{-1} \left(\mathbf{D} - \mathcal{H}(\mathbf{Z}^\mathrm{f})\right),
\end{equation}
where $\mathbf{Y}$ represent the normalized anomaly matrix of the predicted observations $\mathcal{H}(\mathbf{Z}^\mathrm{f})$:
\begin{equation}
\mathbf{Y} = \frac{1}{N-1} \left(\mathcal{H}(\mathbf{z}_1^\mathrm{f})-\overline{\mathcal{H}(\mathbf{Z}^\mathrm{f})}, \ldots,  \mathcal{H}(\mathbf{z}_N^\mathrm{f})-\overline{\mathcal{H}(\mathbf{Z}^\mathrm{f})}\right),
\end{equation}
with
\begin{equation}
\overline{\mathcal{H}(\mathbf{Z}^\mathrm{f})} =\frac{1}{N} \sum_{i=1}^N \mathcal{H}(\mathbf{z}_i^\mathrm{f}). 
\end{equation}
Therefore, the weight matrix $\mathbf{W}$ is solely derived from the observations and the predicted observations, utilizing only the information from the observation space.

In applying an equivariant neural operator for the update, denoted as $\mathcal{F}_\mathrm{E}: \mathbf{Z}^\mathrm{f} \mapsto \mathbf{Z}^\mathrm{a}$, the neural operator $\mathcal{F}_\mathrm{E}$ also adheres to these two principles. Thus, we have
\begin{equation}
    \mathcal{F}_\mathrm{E}: \mathbf{z}_{:,j}^\mathrm{f} \mapsto \mathbf{z}_{:,j}^\mathrm{a} \quad \text{for} \quad j \in \{1, \ldots, n\}.
\end{equation}
The operator $\mathcal{F}_\mathrm{E}$ exclusively depends on the observations $\{\mathbf{d}_i\}_{i=1}^N$ and the predicted observations $\{\mathcal{H}(\mathbf{z}_i^\mathrm{f})\}_{i=1}^N$. Consequently, the ensemble neural filter defines a permutation equivariant mapping:
\begin{equation}
    \mathcal{F}_\mathrm{E}: \left\{\left(z_{i, j}^{\text{f}}, \mathcal{H} (\mathbf{z}_i^\text{f}), \mathbf{d}_i\right)\right\}_{i=1}^N \mapsto \left\{z_{i, j}^{\text{a}}\right\}_{i=1}^N
\end{equation}
for all $j \in \{1, \ldots, n\}$.

\end{document}